\documentclass[journal]{IEEEtran}
\usepackage{amsmath,amssymb,amsfonts}
\usepackage{graphicx}
\usepackage{booktabs}
\usepackage{textcomp}
\usepackage{xcolor}
\usepackage{amsthm}
\usepackage{indentfirst} 
\usepackage{siunitx}
\usepackage{subcaption}
\usepackage{multirow}
\usepackage{float} 
\usepackage{amssymb}
\usepackage{algorithm}
\usepackage{textcase}
\usepackage[noend]{algpseudocode}
\usepackage{hyperref}
\hypersetup{
    colorlinks=true, 
    linkcolor=blue,  
    filecolor=blue,  
    urlcolor=blue,   
    citecolor=blue   
}

\newtheoremstyle{bolddefinition}
  {}
  {}
  {}
  {}
  {\bfseries}
  {.}
  { }
  {}

\theoremstyle{bolddefinition}
\newtheorem{definition}{\textbf{\textit{Definition}}}
\usepackage{enumitem} 

\ifCLASSINFOpdf

\else

\fi

\begin{document}

\title{Agile Decision-Making and Safety-Critical Motion Planning for Emergency Autonomous Vehicles}

\author{Yiming Shu, Jingyuan Zhou, Fu Zhang\thanks{Yiming Shu and Fu Zhang are with the University of Hong Kong, Department of Mechanical Engineering, Hong Kong SAR. 999077, China. Email:{\ yiming.shu@connect.hku.hk , fuzhang@hku.hk} 

Jingyuan Zhou is with the National University of Singapore, Department of Civil and Environmental Engineering, Singapore  119077. Email: jingyuanzhou@u.nus.edu}}

\markboth{IEEE Transactions on Intelligent Transportation Systems}%
{Shell \MakeLowercase{\textit{et al.}}: Bare Demo of IEEEtran.cls for IEEE Journals}

\maketitle

\begin{abstract}
Efficiency is critical for autonomous vehicles (AVs), especially emergency AVs. However, most existing methods focus on regular vehicles, overlooking the different strategies required by emergency vehicles to address the challenge of maximizing efficiency while ensuring safety. In this paper, we propose an Integrated Agile Decision-Making with Active and Safety-Critical Motion Planning System (IDEAM). IDEAM focuses on enabling emergency AVs, such as ambulances, to actively achieve efficiency in dense traffic scenarios with safety in mind. Firstly, the speed-centric decision-making algorithm named the long short-term spatio-temporal graph-centric decision-making (LSGM) is given. LSGM comprises conditional depth-first search (C-DFS) for multiple path generation as well as methods for speed gains and risk evaluation for path selection, which presents a robust algorithm for high efficiency and safety consideration. Secondly, with an output path from LSGM, the motion planner reconsiders environmental conditions to decide constraint states for the final planning stage, among which the lane-probing state is designed for actively attaining spatial and speed advantage. Thirdly, under the Frenet-based model predictive control (MPC) framework with final constraints state and selected path, the safety-critical motion planner employs decoupled discrete control barrier functions (DCBFs) and linearized discrete-time high-order control barrier functions (DHOCBFs) to model the constraints associated with different driving behaviors, making the optimization problem convex. Finally, we extensively validate our system using scenarios from a randomly synthetic dataset, which reveal that IDEAM improves average route progress by approximately $5.25\%$ to $12.93\%$ and increases average speed by about $4.5\%$ to $9.8\%$ compared to the benchmark method, demonstrating its capability to achieve speed benefits and assure safety simultaneously.


Simulation video is available at: \url{https://www.youtube.com/watch?v=873BZoQSf-Q}

Our implementation code is available at \url{https://github.com/YimingShu-teay/IDEAM.git}.

\end{abstract}
\section{Introduction}

With the advancement of perception and communication technologies, autonomous vehicles (AVs) are increasingly integrating into urban and highway traffic environments. In such highly interactive scenarios, it is anticipated that AVs have the potential to enhance driving safety~\cite{mullakkal2020comparative,zheng2022safe} and efficiency~\cite{li2020efficent,mahajan2024improving,wang2024iterative}, which are particularly critical for emergency AVs (EAVs). EAVs refer to vehicles designed to handle urgent, time-sensitive tasks such as ambulances, fire trucks, and emergency service auto taxis ~\cite{shu2023safety}. Existing studies predominantly focus on decision-making \cite{fu2020decision,tian2018adaptive,zheng2021behavioral} or motion planning \cite{ma2015efficient,xie2022distributed,wang2019crash}, respectively, and others propose integrated systems \cite{hang2020integrated,wu2023integrated} designed for regular AVs in urban or highway environments, considering efficiency and safety. However, the applications of AVs assigned to special tasks like emergency vehicles are rarely explored. Therefore, it is crucial to design a system that integrates decision-making and motion planning to maximize efficiency gains while maintaining safety. In this paper, we address these gaps by developing an integrated system specifically tailored for emergency AVs. Our work advances in decision-making and motion planning, providing specialized solutions that prioritize high-speed performance.


\begin{figure}[t]
\centerline{\includegraphics[width=0.47\textwidth,height=0.43\textwidth]{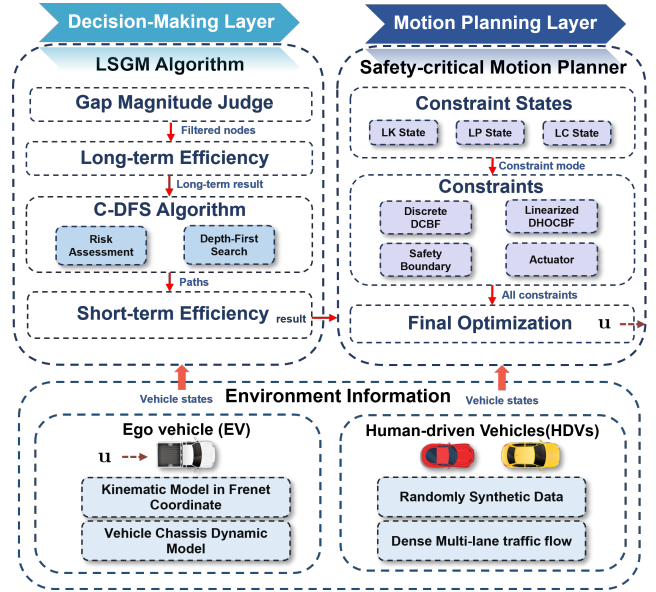}}
\caption{The proposed IDEAM framework. The LSGM decision-making layer is utilized to generate the graph path with the highest efficiency and the final desired vehicle group, while the motion planner actively and safely navigates the AV to the desired vehicle group.}
\label{framework}
\vspace{-0.3cm}
\end{figure}

Decision-making and motion planning are critical for autonomous vehicles, especially in dynamic traffic environments. Learning-based methods for decision-making, such as SVMs~\cite{vallon2017machine}, CNNs~\cite{zhang2022learning}, have been explored for lane-changing decisions. However, deploying such methods in EAVs may present challenges, as specialized datasets for priority-driven navigation are scarce, and their collection may pose difficulties. On the other hand, rule-based approaches, for example, FSM frameworks~\cite{shu2023safety}, offer interpretability but often struggle with scalability. For motion planning, optimization-based approaches using model predictive control (MPC)~\cite{yin2022distributed,zheng2024spatiotemporal,wu2022route,kabzan2019learning} augmented with control barrier functions~\cite{zhu2019barrier} have emerged as a promising safety-critical solution in recent years. However, in dynamic settings, the implicit feasible region in nonlinear programming problems may lead to the risk of local optima~\cite{zeng2021safety}. These challenges are exacerbated for emergency vehicles requiring rapid, efficient, and safety-critical responses.

To address the above-mentioned issues, we propose an Integrated Agile Decision-Making with Active and Safety-Critical Motion Planning System (IDEAM). As shown in Fig. \ref{framework}, our proposed system consists of a decision-making layer and a motion-planning layer. The decision-making is generated by the LSGM algorithm, which is an agile and efficiency-oriented method with strong scalability. The LSGM algorithm solves a graph-based path-finding problem by maximizing speed gains while ensuring safety. Using the C-DFS strategy, the algorithm efficiently identifies paths within the graph, leveraging both short-term and long-term speed benefits. Additionally, by incorporating spatio-temporal risk assessment and gap magnitude judgment, the LSGM algorithm ensures the safety of the selected paths. Ultimately, it provides an optimized and safe path to the motion planning layer. For the motion planning layer, we integrate various constraint states within the MPC framework to formulate different optimization problems, among which the lane-probing constraint state enables AVs to actively explore spatial advantages rather than merely adopting a passive safety mode in response to the environment. Moreover, corresponding to different constraint states, we apply distinct decoupled DCBF and dynamically linearized DHOCBF under the framework, making the optimization problems under these states convex. In summary, the main contributions of this paper are as follows:
\begin{itemize}
\item We design an efficiency-oriented LSGM algorithm with safety in mind that combines spatio-temporal integration and agility, capable of generating reliable decisions that achieve high-speed gains.
\item We introduce a safety-critical motion planner capable of dynamically adjusting constraints based on varying constraint states and actively exploring spatial advantages. The entire motion planning process is framed as a convex numerical optimization problem.
\item The proposed IDEAM system has been rigorously tested in numerous scenarios while compared with several baselines, demonstrating its robust capability for speed benefits and safety assurance.
\end{itemize}

The rest of this paper is organized as follows. Section~\ref{sec:related_works} reviews relevant work on decision-making and motion planning. Section~\ref{sec:Background} presents the preliminary of this paper. Section~\ref{sec:decision} introduces the LSGM algorithm. The motion planning framework is detailed in Section~\ref{sec:motion planning}. Section~\ref{sec:experinment} discusses the simulation results of the proposed method. Section~\ref{sec:discussion} analyzes the system’s limitations and potential directions for future work. Section~\ref{sec:conclusion} concludes the paper.


\section{Related Works}
\label{sec:related_works}
\textbf{\textit{Decision Making}} Recently, various decision-making methods have been developed for autonomous vehicles to handle complex interactive scenarios \cite{crosato2024social}. Traditional traffic models include Minimizing Overall Braking Induced by Lane changes (MOBIL) \cite{kesting2007general,moghadam2021autonomous}, which improves overall traffic efficiency by minimizing braking impacts during lane changes. However, such models lack the emphasis on pursuing speed benefits, which is essential for EAVs. Rule-based methods, such as Finite State Machine (FSM) \cite{chen2020hierarchical}, \cite{shu2023safety}, explicitly define vehicle behaviors through predefined states and transitions, but their complexity rapidly increases as the number of interacting agents grows, leading to large and complex state tables with limited flexibility. Additionally, Partially Observable Markov Decision Processes (POMDP) \cite{gonzalez2019human,ulfsjoo2022integrating} and game-theoretic methods \cite{sun2020game,zhang2024automated}, are regarded as theoretically well-founded decision-making methods considering interaction and environmental uncertainties among vehicles. However, as the number of interacting vehicles increases, the state or action space may grow combinatorially, raising concerns about computational complexity. To address this, recent works \cite{li2023game,huang2023gameformer}, \cite{niu2024planning} have incorporated learning-based trajectory prediction \cite{wong2024socialcircle}, \cite{wong2022view}, \cite{wong2023msn}, \cite{xia2022cscnet} to replace the behavior simulations. The incorporation of data-driven components into decision-making relies on large-scale datasets. However, most existing datasets are tailored to common urban civilian driving scenarios, limiting the generalizability of such methods to more specialized or high-priority contexts. This challenge similarly affects learning-based decision-making approaches, such as those utilizing CNNs \cite{zhang2022learning} and SVMs \cite{liu2019novel}, \cite{vallon2017machine}, \cite{benterki2019prediction}. In contrast, reinforcement learning (RL)-based methods \cite{jia2024learning}, \cite{mo2019decision,yavas2020new} enable policy optimization through interaction with simulated environments without relying on labeled datasets, but it typically consider safety only through reward functions, which may be insufficient to address collision risks during exploration.
 
\textbf{\textit{Motion Planning}} Applications in lane keeping \cite{ding2024adaptive,ding2022security}, lane changing \cite{mehmood2019trajectory, hu2022adaptive}, and merging \cite{zhang2024automated} have become key research directions in motion planning for autonomous vehicles in recent years. Among various approaches, control barrier functions (CBFs) have emerged as a promising choice, offering safety guarantees via forward invariance to keep the system within predefined safe sets.
A CLF-CBF-QP framework was first proposed by \cite{ames2014control} for adaptive cruise control, and was later extended by \cite{he2021rule}, who integrated it with a rule-based switching strategy to enable safe autonomous lane changes. However, the method provides only one-step safety guarantees without considering long-term planning horizons. Model predictive control (MPC), on the other hand, optimizes over a receding horizon, enabling robust trajectory planning with explicit handling of constraints. \cite{zeng2021safety} first introduced the integration of MPC with discrete-time control barrier functions (MPC-DCBF), where the feasibility of the resulting nonlinear programming problem (NLP) was analyzed, highlighting the tradeoff between safety and feasibility. Further applications of MPC-DCBF include \cite{he2022autonomous}, which demonstrated its effectiveness in low-speed car racing scenarios. Recently, the linearized discrete high-order control barrier function (DHOCBF) has emerged as an effective solution to the nonlinear programming (NLP) problem, offering fast computational performance while ensuring safety. While it has been applied to static obstacle avoidance, it holds promise for extension to dynamic interactive scenarios.

\section{Preliminary}
\label{sec:Background}

\begin{figure}[t]
\centerline{\includegraphics[width=0.39\textwidth,height=0.26\textwidth]{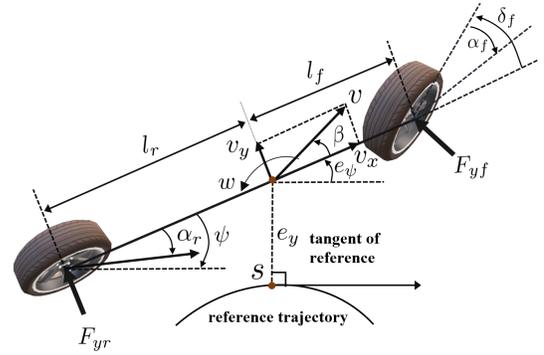}}
\caption{Diagram of vehicle dynamics model.}
\label{dynamics}
\vspace{-0.3cm}
\end{figure}

In this section, the preliminary related to the motion planning module is introduced. We present the vehicle dynamics model in Section \ref{subsec: Vehicle Dynamics Model}, while some background of the control barrier function is introduced in Section \ref{subsec:DHOCBF}.

\subsection{Vehicle Dynamics Model}
This subsection first introduces a precise description of the vehicle dynamics model, which includes the kinematic model in Frenet coordinates and the vehicle chassis dynamic model. Subsequently, the linearized model for state constraint and state propagation is introduced.

\label{subsec: Vehicle Dynamics Model}
\noindent\textbf{\textit{Kinematic Model in Frenet Coordinate.}} The kinematic model within the Frenet frame models the motion of vehicles along a reference trajectory. The Frenet kinematics is described as follows:
\begin{align}
&\dot{s} = \frac{v_x \cos{(e_{\psi})} - v_y \sin{(e_{\psi})}}{1- \kappa \,(s)e_y},\\
&\dot{e}_{\psi} = \dot{\psi} - \frac{v_x \cos{(e_{\psi}) - v_y \sin{(e_{\psi}})}}{1-\kappa \, (s) e_y} \kappa \, (s),\\
&\dot{e}_y= v_x \sin {(e_{\psi})} + v_y \cos{(e_{\psi})},
\end{align}
where $s$ represents the distance traveled along the lane's centerline, $e_y$ and $e_{\psi}$ denote the deviation distance and the heading angle error between the vehicle and the lane's centerline, respectively. Besides, $\kappa$ refers to the curvature at the reference point.

\noindent\textbf{\textit{Vehicle Chassis Dynamic Model.}} The vehicle chassis dynamic model integrates multidirectional motions and tire forces to construct a comprehensive vehicle dynamics model. The dynamics model shown in Fig.~\ref{dynamics} are given by:
\begin{align}
&\dot{v}_x = a_x -\frac{1}{m} F_{fy} \sin{(\delta_f)} + w v_y,\\
&\dot{v}_y = \frac{1}{m}(F_{fy} \cos{(\delta_f)} + F_{ry}) - w v_x,\\
&\dot{w} = \frac{1}{I_z}(l_f F_{fy} \cos{(\delta_f)} - l_r F_{ry}),
\end{align}
where $v_x$ and $v_y$ denote the longitudinal velocity and lateral velocity, respectively. $w$ represents the yaw rate of the vehicle. $a_x$ is the longitudinal acceleration and $\delta_f$ denotes steer angle. $F_{fy}$ and $F_{ry}$ are the lateral forces on the front and rear tires, while $m$ stands for the mass. $I_z$ indicates the moment of inertia. $l_f$ and $l_r$ refer to the distances from the center of the vehicle to the front and rear axles, respectively.

Pacejka's Magic Formula \cite{pacejka2005tire} is adopted to model the lateral forces of the tires as follows
\begin{align}
&F_{fy} = 2 D_f \sin{({C_f \arctan{(B_f \alpha_f)}})}\\
&F_{ry} = 2 D_r \sin{({C_r \arctan{(B_r \alpha_r)}})}
\end{align}
where $B$, $C$ and $D$ are model coefficients. The sideslip angles for the front and rear tires are denoted by $\alpha_f$ and $\alpha_r$. Specifically, they are formulated as follows
\begin{align}
&\alpha_f = \delta_f - \arctan{(\frac{l_f w+v_y}{v_x})}\\
&\alpha_r =\arctan{(\frac{l_r w+v_y}{v_x})}
\end{align}


\noindent\textbf{\textit{Model Linearization.}} For simplicity of computation, we then linearize the above-mentioned vehicle dynamics at its operating point. As demonstrated in \cite{sakai2018pythonrobotics}, we use the trajectory optimized in the last step for linearization and employ the Euler method to derive a discretized model:
\begin{equation}
    \mathbf{x}_{k+1} = \mathbf{A} \mathbf{x}_k +\mathbf{B}\mathbf{u}_k +\mathbf{C}
\end{equation}
where $\mathbf{A}$, $\mathbf{B}$ and $\mathbf{C}$ are system matrices. $\mathbf{x}$ and $\mathbf{u}$ are the state vector and input vector formulated as 
\begin{equation}
\mathbf{x}= \begin{bmatrix}
    v_x, v_y, w, s, e_y, e_{\psi} 
\end{bmatrix},  \mathbf{u}= \begin{bmatrix} a_x, \delta
\end{bmatrix}
\end{equation}

\subsection{Control Barrier Functions}
We then introduce the discrete control barrier function that enforces the safety of the affine control system:
\begin{equation}
    \dot{\mathbf{x}} = f(\mathbf{x}) + g(\mathbf{x})\mathbf{u}
    \label{affine}
\end{equation}
where $\mathbf{x} \in \mathbb{R}^{n}$, $f: \mathbb{R}^{n} \rightarrow \mathbb{R}^{n}$ and $g: \mathbb{R}^{n} \rightarrow \mathbb{R}^{n\times q}$ are locally Lipschitz, $\mathbf{u} \subset \mathbb{R}^{q}$.

\begin{definition}[Discrete Control Barrier Function \cite{zeng2021safety}] Let $\mathcal{C}= \left\{ \mathbf{x} \in \mathbb{R}^n : h(\mathbf{x},t) \geq 0 \right\}$ be the superlevel set of the continuous differentiable function $h:D\subset \mathbb{R}^n \rightarrow \mathbb{R}$, then $h$ is a control barrier function (CBF) for system \eqref{affine} if $\frac{\partial h}{\partial \mathbf{x}}(\mathbf{x}) \neq 0$ for all $\mathbf{x} \in \partial \mathcal{C}$ and there exists an class $\mathcal{K}$ functions $\gamma$ such that:
\begin{equation}
   \exists \mathbf{u} \text { s.t. } \dot{h}(\mathbf{x},\mathbf{u}) \geq - \gamma h(\mathbf{x})
    \label{eq: cbf}
\end{equation}

When referring to the discrete-time domain, the discrete CBF (DCBF) formulation~\eqref{eq: cbf} can be shown as follows
\begin{equation}
    \Delta h(\mathbf{x}_k,\mathbf{u}_k) \geq -\gamma h(\mathbf{x}_k), 0 < \gamma \leq 1
    \label{dcbf}
\end{equation}
where $\Delta h(\mathbf{x}_k,\mathbf{u}_k) := h(\mathbf{x}_{k+1})-h(\mathbf{x}_k)$.
\end{definition}

In the context of our study, the relative degree \cite{xiao2019control} of the continuously differentiable function $h: \mathbb{R}^n \to \mathbb{R}$ is higher than 1 concerning the system (\ref{affine}). Consequently, we introduce the discrete-time high-order control barrier function (DHOCBF) approach to address this characteristic.

\begin{definition}[Discrete-time High-Order Control Barrier Function \cite{xiao2019control,liu2023iterative}]
For a $m$-th order differentiable function $h: \mathbb{R}^n \times [t_0,\infty) \to \mathbb{R}$, a series functions $\psi_0:\mathbb{R}^n \times [t_0,\infty) \to \mathbb{R}, \psi_1:\mathbb{R}^n \times [t_0,\infty) \to \mathbb{R},\cdots, \psi_m:\mathbb{R}^n \times [t_0,\infty) \to \mathbb{R}$ are defined
 in the form:
\begin{equation}
\begin{aligned}
\psi_0(\mathbf{x},t) &:= h(\mathbf{x},t) \\
\label{dhocbf1}
&\ \vdots  \\
\psi_m(\mathbf{x},t) &:= \Delta \psi_{m-1}(\mathbf{x},t) + \alpha_m(\psi_{m-1}(\mathbf{x},t)),
\end{aligned}
\end{equation}
where $\alpha_1(\cdot), \alpha_2(\cdot), \ldots, \alpha_m(\cdot)$ denote class $\mathcal{K}$ functions.

A series of sets $C_1(t),C_2(t),...,C_m(t)$ are further defined as follows:
\begin{align}
C_1(t) &:= \{\mathbf{x} \in \mathbb{R}^n : \psi_0(\mathbf{x},t) \geq 0\} \nonumber \\
\label{DHOCBF2}
&\ \vdots \\
C_m(t) &:= \{\mathbf{x} \in \mathbb{R}^n : \psi_{m-1}(\mathbf{x},t) \geq 0\} \nonumber
\end{align}

Let \(\psi_i(\mathbf{x}, t)\), \(i \in \{1, \ldots, m\}\) be defined by \eqref{dhocbf1} and \(C_i\), \(i \in \{0, \ldots, m-1\}\) be defined by \eqref{DHOCBF2}. A function \(h : \mathbb{R}^n \to \mathbb{R}\) is a DHOCBF with relative degree \(m\) for system (\ref{affine}) if there exist \(\psi_m(\mathbf{x}, t)\) and \(C_i\) such that
\[
\psi_m(\mathbf{x}, t) \geq 0, \quad \forall \mathbf{x} \in C_1 \cap \cdots \cap C_{m}.
\]
\end{definition}

DHOCBF typically divides the space into feasible and infeasible regions, with the feasible region being non-convex. To improve computational efficiency and the quality of solutions \cite{liu2023iterative}, these constraints can be convexified by linearizing DHOCBF in the temporal domain.

\noindent\textbf{\textit{Linearized DHOCBF}}. 
The linearization of DHOCBF is achieved by projecting the ego vehicle's trajectory onto the nearest boundary of the obstacle for each future time step, as depicted in Fig.~\ref{cbf}. Employing the linearization technique specified in \cite{liu2023iterative}, it is possible to linearize the DHOCBF to its highest order, culminating in the derivation of universally reformulated convex constraints:
\vspace{-0.05cm}
\begin{equation}
\begin{split}
\psi_{i-1}(\mathbf{x}_{t,k}) + \sum_{\nu=1}^{i} Z_{\nu,i} (1 - \gamma_i)^k \psi_0(\mathbf{x}_{t,\nu}) &\geq \\
\omega_{t,k,i} Z_{0,i} (1 - \gamma_i)^k \psi_0(\mathbf{x}_{t,0})
\label{dhocbfexplan}
\end{split}
\end{equation}

for $i \in \{1, \ldots, m_{cbf}\}$, $\omega_{t,k,i} \in \mathbb{R}.$
where $Z_{\nu, i}$ is a constant given $\nu \in \{0, \ldots, i\}$,  Here, $i$ represents the order of the DCBF, which dictates the number of derivatives considered, and $m_{cbf}$ denotes the highest order of the DCBF. Meanwhile, $k$ represents the time step index in the prediction horizon. 

For collision avoidance problems, DCBFs with $m_{cbf}=2$ are typically used. Hence, we pay particular attention to the form of the constraints (first-order and second-order constraints) after the DCBFs have undergone linearization:
\vspace{-0.25cm}

\begin{equation}
\begin{aligned}
    \psi_{0}(x_{t,k+1}) &\geq w_{t,k,1}(1-\gamma_1)^{(k+1)}\psi_{0}(x_{t,0}), \\
    \psi_{1}(x_{t,k+1}) &\geq (1-\gamma_2)^{(k+1)}\psi_{0}(x_{t,1}) \\
    & + w_{t,k+1,2}(\gamma_1-1)(1-\gamma_2)^{(k+1)}\psi_{0}(x_{t,0})
    \label{m2dhocbf}
\end{aligned}
\end{equation}
where  $k\in(0,....N-1) $ for the first order constraints and $k\in(0,....N-2) $ for the second order constraints.

\section{Long short-term spatio-temporal graphcentric decision-making}
\label{sec:decision}
This section introduces the LSGM algorithm. With the information about nearby vehicles, the LSGM can make lane-changing decisions for the ego vehicle.
\begin{figure}[t]
\centerline{\includegraphics[width=0.49\textwidth,height=0.245\textwidth]{cbf.pdf}}
\caption{Visualization of linearized DHOCBF. The linearized safe set, shown in blue, guarantees collision avoidance. {The gray area represents the infeasible region segmented by DHOCBF, while the green area denotes the infeasible region delineated by the linearized DHOCBF at time step $k$.}}
\label{cbf}
\end{figure}
\label{subsec:DHOCBF}
\subsection{Problem Formulation}
In dense multilane traffic with strong interactions, lane-changing tasks can be viewed as finding a path composed of vehicle groups that comprise the preceding and following vehicles, offering higher efficiency gains, greater safety, and flexible space. Given that these vehicle groups densely populate the traffic flow, which itself is subject to frequent changes, the crucial challenge for the decision-making layer is to secure long-term and stable benefits in terms of speed, safety, and space in the future spatio-temporal context.

To address this critical issue in lane-changing decisions, the LSGM algorithm not only integrates the evaluation of both long-term and short-term efficiency gains but also incorporates gap magnitude judgment and risk assessment to ensure sufficient safety and space for vehicle maneuvers. Within LSGM, we have designed a conditional depth-first search (C-DFS) algorithm that embeds risk assessment to search for connected paths composed of vehicle groups. To be specific, within each lane, two vehicle groups formulated by two leading vehicles and one following vehicle are considered: in the left lane, these groups are referred to as $L_1$ and $L_2$; in the right lane as $R_1$ and $R_2$; and in the middle lane as $C_1$ and $C_2$. 
Before evaluating long-term and short-term efficiency and performing the C-DFS algorithm, the gap magnitude judge identifies nodes without sufficient space, which are then excluded from consideration. This step is crucial as ensuring a sufficient gap space allows vehicles to maneuver flexibly and provides essential reaction space to respond to emergencies.

Generally, with the graph filtered by gap magnitude judgment, the ego vehicle will originate from one of these six vehicle groups, which then becomes the starting node for the C-DFS, while the end node, representing the vehicle group with the highest long-term efficiency gain, is selected by the Long-Term Efficiency Group Selector. After inputting the start and end nodes, the C-DFS algorithm rapidly generates multiple valid connected paths like $C_1 \rightarrow L_2 \rightarrow C_2 \rightarrow R_2$, among which the Short-Term Efficiency Group Selector chooses the path with the highest short-term speed benefit.

\begin{figure}[t]
\centerline{\includegraphics[width=0.48\textwidth,height=0.27\textwidth]{risk.pdf}}
\caption {An illustration of risk assessment within the decision-making time domain in the C-DFS algorithm. A node consists of three components: the leading vehicle, the following vehicle, and the gap between them (represented by the gray area in the figure). When C-DFS determines the first node to visit, the time horizon is truncated to the first segment, i.e., $z=1$. Here, $k$ represents the timestep within each time segment $z$. The lower part of the figure illustrates that at $k=2$ in the $z=1$ time segment, the assessment of $d_{\text{risk}}$ is based on the leader vehicle of node $C_1$ and the following vehicle of node $R_1$.
}

\label{risk}
\end{figure}

\vspace{-0.08cm}
\subsection{Conditional Depth-First Search}
\label{sec:C-DFS}
The C-DFS algorithm is designed to efficiently search for valid paths in multi-lane scenarios while incorporating safety considerations through integrated risk assessment. This section introduces the key components and procedures of the C-DFS algorithm, which ensures that only safe and feasible paths are explored and selected.


\noindent\textbf{\textit{Risk Assessment.}} When searching for connected paths, the safety of connectivity between nodes is crucial for successful and accident-free lane changes. Risk assessment is employed to determine the safety of the connection between adjacent nodes within a path. If this connection is deemed unsafe, the path is subsequently filtered out. As depicted in Fig.~\ref{risk}, ensuring a safe connection between two nodes necessitates a sufficient temporal gap $d_{risk}$ between the lead vehicle of the pointing node and the rear vehicle of the pointed node, thus enabling the ego vehicle to pass through securely. During the risk assessment process, when evaluating the connection between each pair of nodes, distinct time domains in the prediction horizon are considered, each characterized by a fixed length. Earlier connections correspond to preceding time domains, while subsequent connections correspond to succeeding time domains. The mathematical formulation of $d_{risk}$ is defined as follows
\begin{equation}
    d_{risk}^{z,k}=\begin{cases}2l_{diag}+\epsilon & v_{l,\text{ego}
    }^{z,k}>v_{f,i}^{z,k}\\2l_{diag}+n\Delta v +\epsilon & v_{f,i}^{z,k}> v_{l,\text{ego}}^{z,k}\end{cases}
\end{equation}
where $\Delta v = v_{l,\text{ego}}^{z,k} - v_{f,i}^{z,k}$, $z_d \subseteq \{z_1,z_2,z_3...,z_{N_{seg}}\}$, $k \in z_d$, $d$ represents the depth in the C-DFS algorithm. Here, $N_{seg}$ is the number of a certain time domain segment, and $l_{diag}$ denotes the diagonal length of the vehicle. Besides, $n$ and $\epsilon$ are constant positive parameters. The index $i$ refers to the group currently being evaluated.

\noindent\textbf{\textit{Conditional DFS.}} Risk assessment will be integrated into C-DFS for path finding as safety criteria. This subsection introduces the C-DFS algorithm as detailed in Algorithm \ref{alg:find_all_paths}, a core method in the LSGM algorithm for recursively finding valid paths. The C-DFS algorithm comprises the following functions:
\begin{itemize}
\item $\mathbf{UpdateMaxLevelVisited}$: record the highest group level accessed for each lane. For every lane, we consider two vehicle groups, group $1$ and group $2$, where group $2$ occupies a more advantageous spatial position and also possesses a higher level.
\item $\mathbf{LevelCheck}$: verify if a node's level is below the highest recorded level. Should it be lower, the search for paths passing through this node will be disregarded, an example is presented in Fig.~\ref{decision1}. This approach ensures that once a path progresses through the spatially advantageous group $2$, it does not revert to group $1$.
\item $\mathbf{RiskAssessment}$: for each node, during this recursive iteration, evaluate at every step within the time segment of the prediction at depth $d$, ensuring that the safety standards are met between the follower vehicle associated with the pointed node and the leader vehicle of the pointing node.

\item \textbf{Complexity Analysis}: A standard DFS performs an exhaustive search with a worst-case complexity of $O(b^d)$. In contrast, C-DFS leverages pruning mechanisms, including LevelCheck and RiskAssessment, and benefits from a limited number of candidate nodes per step, which together substantially reduce the search space. Consequently, the practical complexity is $O(\min(P, b^d, N+E))$, where $P$ is the number of valid paths, $b$ is the branching number, $d$ is the search depth, and $N$ and $E$ denote the number of nodes and edges, respectively. Empirically, C-DFS achieves near-constant time performance with negligible computational cost, supporting its suitability for real-time applications.
\end{itemize}

\begin{figure}[t]
\centerline{\includegraphics[width=0.37\textwidth,height=0.18\textwidth]{levelcheck.pdf}}
\vspace{0.3cm}
\caption{n illustration of path generation and LevelCheck in C-DFS. When the start and end nodes are determined, the C-DFS searches for multiple safe paths validated through risk assessment. Taking the path from the start node $L_1$ to the end node $L_2$ as an example, the figure illustrates multiple possible paths from $L_1$ to $L_2$. Importantly, LevelCheck prevents paths from reverting from group $2$ to group $1$ by filtering out nodes with lower levels, for example, paths will not revert to $R_1$ or $C_1$ once they progress through $R_2$ or $C_2$. }
\label{decision1}
\end{figure}

\begin{algorithm}
\caption{C-DFS}\label{alg:find_all_paths}
\begin{algorithmic}[1] 
\State \textbf{Input:} input graph connections $\mathcal{G}_{in}$, input nodes $\mathcal{N}_{in}$
\State \ \ \ \ \ \ \ \ \ start node $n_s$, end node $n_e$, depth $d$
\State \ \ \ \ \ \ \ \ \ current traversal path $p_c$, visited level record $\mathcal{V}$
\State \textbf{Output:} valid paths $\mathcal{P}_{\text{all}}$ 
\State \textbf{C-DFS Algorithm}
\State add $n_s$ to $p_c$
\State $\mathcal{V} \gets \text{UpdateMaxLevelVisited($n_s$)}$
\If{$n_s == n_e$}
\State \Return $p_c$
\EndIf
\State set $\mathcal{P}_{\text{all}}$ empty
\For{each node \(n\) in $\mathcal{G}_{in}(n_s)$}
    \If {LevelCheck($n$, $\mathcal{V}$)}
        \State continue
    \EndIf
    \If{ not RiskAssessment($\mathcal{G}_{in}(n_s)$,$n_s$,$n$,$d$)}
        \State continue 
    \EndIf
    \State $\mathcal{P_{\text{new}}} =$ \text{C-DFS}$(\mathcal{N}_{in}, \mathcal{G}_{in},p_c,\mathcal{V},n,n_e, d=d+1)$
    \For{each path $p$ in $\mathcal{P_{\text{new}}}$}
        \State add \(p\) to \(\mathcal{P}_{\text{all}}\)
    \EndFor
\EndFor
\State \Return $\mathcal{P}_{\text{all}}$
\end{algorithmic}
\end{algorithm}

\subsection{Long Short-term Efficiency Group Selector}
Much works focus on improving efficiency by following a faster leader vehicle. However, considering only the speed of the leader vehicle is insufficient, as larger spaces in other vehicle groups may provide opportunities for overtaking and other maneuvers beneficial for speed gains. Unlike these methods, we refine the assessment by evaluating the predicted spatial positions of surrounding vehicles. The longitudinal position of the leader vehicle in future time frames not only implies information about the vehicle's speed but also indicates the availability of sufficient space for maneuvers such as overtaking, accelerating, and lane-changing, all aimed at enhancing efficiency. Inspired by this notion, we categorize the efficiency evaluation of a path into two aspects: long-term efficiency and short-term efficiency. 

\noindent\textbf{\textit{Long-term Efficiency}}. 
The evaluation of long-term efficiency is notably straightforward. Within an extended prediction period $T_{\text{long}}$, the gap group containing the leader vehicle positioned farthest in the longitudinal space is selected as the graph's long-term target node. 

\noindent\textbf{\textit{Short-term Efficiency}}.
When there are multiple viable paths, the path with the better short-term efficiency will be chosen as the final path. Similar to the selection of a long-term target node, the assessment of short-term efficiency is based on comparing the leader vehicle's longitudinal position within a notably shorter prediction period $T_{\text{short}}$. The short-term target node is the node closest to the starting node along a path, representing the next vehicle group to be approached. Unlike the evaluation of long-term efficiency, to prevent decision instability caused by minimal longitudinal distance differences between leader vehicles of the nearest nodes in multiple paths, the comparison process employs progressively longer prediction periods if those distances are below a specified threshold $d_{threshold}$. This gradual extension of the prediction period from an initial short-term horizon $T_{\text{short}}$ continues methodically until it aligns with the long-term forecast duration $T_{\text{long}}$, thereby ensuring decisions are made on a more stable and reliable basis when faced with closely matched conditions.
\vspace{-0.06cm}
\subsection{LSGM Framework}
This subsection summarizes the proposed method of long short-term spatio-temporal graph-centric decision-making as detailed in Algorithm \ref{alg:euclid}. The LSGM algorithm runs at every timestep, with Algorithm \ref{alg:euclid} outlining the steps followed during a single timestep. This Algorithm consists of the following functions:
\begin{itemize}
\item $\mathbf{CurrentNode}$: assess the ego vehicle's position and match it to the corresponding node, which should be either $L_1$, $C_1$, or $R_1$, return node $n_{\text{start}}$.
\item $\mathbf{GapMagnitudeJudge}$: evaluate each node for sufficient gap space and return the nodes $N_e$ that lack adequate gap space.
\item $\mathbf{UpdateGraphAndNodes}$: remove all connections related to $N_e$ from the graph and exclude $N_e$ from $N$, return $N_{e^-}$.
\item $\mathbf{LongTermEfficiency}$: identify and return the node $n_{\text{long}}$ that offers the maximum long-term efficiency.
\item $\mathbf{FindShortestPath}$: find the paths from $n_{\text{start}}$ to $n_{\text{long}}$ that contain the fewest number of nodes.

\item $\mathbf{ShortTermEfficiency}$:identify and return the node $n_{\text{short}}$ that offers the maximum short-term efficiency.

\item \textbf{Complexity Analysis}: The LSGM algorithm extends C-DFS by introducing an iterative decision-making process that evaluates long-term efficiency, executes C-DFS, and refines the decision space through graph updates. The main computational cost arises from the while loop, which iteratively performs efficiency evaluation ($O(N)$), C-DFS ($O(\min(P, b^d, N+E))$), and graph updates ($O(N+E)$) until a valid decision is found. In the worst case, with up to $O(N)$ iterations and ineffective pruning, the overall complexity is $O(N(N+E))$. However, empirical results show that pruning significantly reduces the search space, resulting in an effective complexity closer to $O(NP)$.

\vspace{0.3cm}
\end{itemize}
\begin{algorithm}
\caption{LSGM}\label{alg:euclid}
\begin{algorithmic}[1]
\State \textbf{Input:} nodes \ $\mathcal{N}$, $\text{constructed graph connections} \ \mathcal{G}_c$
\State \ \ \ \ \ \ \ \ \ current node $n_{\text{start}}$, depth $d=1$
\State \ \ \ \ \ \ \ \ \ empty traversal path $p_c$, empty level record $\mathcal{V}$
\State \textbf{Output:} target path $n_{\text{start}}$

\State \textbf{Gap Maginitude Judgement}
\State $\mathcal{N}_e \gets \text{GapMagnitudeJudge($\mathcal{N}$)}$
\State $\mathcal{N}_{e^-}, \mathcal{G}_{ce^-} \gets \text{UpdateGraphAndNodes}(\mathcal{N}, \mathcal{N}_e, \mathcal{G}_c)$

\State \textbf{LSGM Decision-Making}
\While{True}
    \State $n_{\text{long}} \gets \text{LongTermEfficiency}(\mathcal{N}_{e^-})$
\If{$n_{\text{long}} == n_{\text{start}}$}
    \State $n_{\text{target}}$ = $n_{\text{start}}$
    \State \Return $n_{\text{target}}$
    \EndIf
    \State $\mathcal{P}_{\text{all}} \gets \text{C-DFS}(\mathcal{N}_{e^-}, \mathcal{G}_{ce^-},p_c,\mathcal{V},n_{\text{long}},n_{\text{start}}, d)$
\If{$\mathcal{P}_{\text{all}}$ \text {is not empty}}
\State $\mathcal{P}_s \gets \text{FindShortestPath}(\mathcal{P}_{\text{all}})$
\If{$|\mathcal{P}_s| == 1$}
    \State $n_{\text{target}} = \mathcal{P}_s.\text{first}$
\Else
    \State $n_{\text{target}}$ $\gets$ $\text{ShortTermEfficiency}(\mathcal{P}_s)$
\EndIf
\State \Return $n_{\text{target}}$
\Else
\State $\mathcal{N}_{e^-}, \mathcal{G}_{ce^-} \gets \text{UpdateGraphAndNodes}(\mathcal{N}_{e^-}, \mathcal{G}_{ce^-})$
\EndIf
\EndWhile

\end{algorithmic}
\end{algorithm}

\section{Motion Planning Framework}
\label{sec:motion planning}
After the LSGM decision-making layer outputs a desired vehicle group, the motion planning layer needs to provide a safe motion planning strategy. This section will detail the constraint states and constraint formulation, concluding with a presentation of the entire optimization framework. 

\subsection{Constraints States}
\label{sec:motion planningA}
Constraint states represent different driving behaviors by defining the specific constraints required for each behavior. In this subsection, three types of constraint states are introduced: the lane-changing state, the lane-keeping state, and the lane-probing state. These states will be utilized to formulate constraints for the optimization problem. 

\noindent\textbf{\textit{Lane-keeping State}}. When the desired vehicle group is the current group, the ego vehicle remains in the current lane. For lane keeping, the required constraints are of two types: longitudinal constraints, for constraining the leading and following vehicles in the current lane longitudinally; and lateral constraints, for constraining vehicles around the ego vehicle within a certain range on adjacent lanes laterally.

\noindent\textbf{\textit{Lane-probing State}}. If the desired vehicle group differs from the current vehicle group, it indicates support for a lane change. When the ego vehicle aims to move to a specific vehicle group, the primary interaction is with the follower in the desired group. The lane-probing state emerges when the ego vehicle has not yet attained a spatial advantage over the follower in the desired group, implying the need for the ego vehicle to proactively explore opportunities for merging into the desired group. The ego vehicle will continue to actively probe forward until it achieves the spatial condition $s_e-s_f^d \geq \frac{l}{2}$,  indicating that a sufficient spatial advantage has been gained.
For lane-probing, the required constraints include longitudinal constraints for the following vehicle and ellipse constraints for the leading vehicle in the current group, in addition to lateral constraints. These ellipse constraints, described as linearized DHOCBF constraints in Section \ref{subsec:DHOCBF}, enable the vehicle to safely probe forward for a spatial advantage while facilitating a smooth transition between lane-probing and lane-changing.

\noindent\textbf{\textit{Lane-changing State}}. In the lane-changing state, beyond the foundation set by lane-probing, additional constraints are applied to both the leader and follower of the desired group. For the leader of the desired group, longitudinal constraints are employed to improve the ego vehicle's adaptability as it merges into the desired group. At the same time, the follower is subjected to ellipse constraints.

\vspace{-0.06cm}
\subsection{Constraints Formulation}
 As mentioned in Section \ref{sec:motion planningA}, different states are associated with specific constraints. This subsection elaborates on the different constraints formulation, including longitudinal constraints, lateral constraints, ellipse constraints, safety boundary constraints, and actuator constraints. 

\noindent\textbf{\textit{Longitudinal Constraints}}.
To ensure the safety of the ego vehicle longitudinally, it is crucial to maintain an appropriate distance from both the leading and following vehicles. Also, during lane-changing, longitudinal constraints on the desired group’s leader help ensure smooth adaptation. To achieve this, we adopt a linear DCBF aimed at defining a safety distance barrier. The function is mathematically formulated as
\begin{equation}
    h_{lon}(\mathbf{x}) = |s-s_i|-T_d v_x - d_0
\end{equation}
where $s_i$ indicates the spacing of the surrounding vehicle with $i \in \{cl, cf, dl\}$. To be specific, $cl$ stands for the leader of the current group, $cf$ represents a follower in the current group, and $dl$ is the leader of the desired group. $T_d$ and $d_0$ denote the time headway and the least distance, respectively. The longitudinal constraints can hence be constructed by \eqref{dcbf} as follows
\begin{equation}
    \Delta h_{lon}^i(\mathbf{x}_k,\mathbf{u}_k) \geq -\gamma_{lon}^i h_{lon}^i(\mathbf{x}_k) - \epsilon_{lon}^i
    \label{dcbflon}
\end{equation}
where $k \in \{0,...,N-1\}$, $\epsilon_{lon}^i$ is the slack variable for longitudinal constraints, which is introduced to facilitate smooth transitions in constraints applied to vehicles across different constraint states. To ensure that the constraint is not too conservative, a large cost related to $\epsilon_{lon}^i$ is added when the constraints are violated.

\noindent\textbf{\textit{Lateral Constraints}}.
The lateral constraints are designed for surrounding HDVs, which are only applicable in adjacent lanes. These constraints are active specifically when HDVs enter the region of interest (ROI) \cite{shu2023safety}, which is ego-centric. ROI is defined as
$[s_{ego}-2 l_{diag},s_{ego}+2 l_{diag}]$.

The lateral barrier function is defined as follows:
\begin{equation}
    h_{lat}^j(\mathbf{x}) = |e_y-e_{y}^{proj}|  - w - d_{lat}, \quad j \in \mathcal{J}
\end{equation}
where $\mathcal{J} = \{j \mid e_{y}^{\text{proj}} \in \text{ROI}\}$
$e_{y}^{proj}$ denotes the deviation distance projected on the track centerline of the ego vehicle. Moreover, $w$ is the width of a vehicle, and $d_{lat}$ is a positive constant. The lateral constraints can then be formed the same way as \eqref{dcbf}:
\begin{equation}
    \Delta h_{lat}^j(\mathbf{x}_k,\mathbf{u}_k) \geq -\gamma_{lat}^j h_{lat}^j(\mathbf{x}_k) - \epsilon_{lat}^j
\end{equation}
where $k \in \{0,...,N-1\}$. Similar to longitudinal constraints, the slack variable $\epsilon_{lat}^j$ for lateral constraints is also incorporated.

\noindent\textbf{\textit{Ellipse Constraint.}}  During lane-changing, we adopt linearized DHOCBF to formulate the constraint of the leader in the current vehicle group or the follower in the desired vehicle group. Different from \cite{liu2023iterative}, which used it for static obstacle avoidance and regarded obstacles as a circle, we apply it to dynamic traffic scenarios and adopt elliptical obstacles in the Frenet coordinate system.

As the obstacle area is defined as an ellipse, the way to find a prior trajectory's projected point is different from the scenario when the obstacle is a circle. Defining the ellipse barrier region as
$\frac{(s-s_o)^2}{a^2} + \frac{(e_y-e_{yo})^2}{b^2} = 1$, our purpose is to minimize the following optimization problem:
\begin{subequations}
    \begin{align}
        & \underset{s,e_y}{\text{min}}
        & & J(s,e_y) = (s-s_b)^2 + (e_y-e_{yb})^2 \\
        & & & \text{s.t.} \quad \frac{(s-s_o)^2}{a^2} + \frac{(e_y-e_{yo})^2}{b^2} = 1,
    \end{align}
    \label{opt1}
\end{subequations}
\noindent where the $s_o$ and $e_{yo}$ represent the traveled distance and lateral error of the considered ellipse center, respectively. $(s_b,e_{yb})$ refers to the initial guess for the boundary of the ellipse.

To solve this problem, we introduce the method of Lagrange multipliers. Consequently, the corresponding Lagrangian function for the optimization problem (\ref{opt1}) is as follows:
\begin{align}
L(s, e_y, \lambda) = (s - s_b)^2 + (e_y - e_{y_b})^2  \notag \\
+ \lambda \left( \frac{(s - s_o)^2}{a^2} + \frac{(e_y - e_{y_o})^2}{b^2}-1 \right)
\end{align}
where $\lambda$ is the lagrange multiplier.
By setting the partial derivatives of $L$ with respect to $s$, $e_y$ and $\lambda$ to zero, 
we can find the critical point ($\Bar{s}$, $\Bar{e_y}$) of $L$.

The linearized DHOCBF is the tangent of the ellipse based on the closest point on the ellipse, the detailed formulation is as follows
\begin{subequations}
    \begin{align}
        \psi_0(\mathbf{x}_{t,k}) &= A_{cbf}s_{t,k} + B_{cbf}e_{y,t,k} + C_{cbf}, \\
        A_{cbf} &= b^2(\bar{s}_{t,k}-s_{o,t,k}), \\
        B_{cbf} &= a^2(\bar{e}_{y,t,k}-e_{yo,t,k}), \\
        C_{cbf} &= b^2(s_{o,t,k}^2 - s_{o,t,k}\bar{s}_{t,k}) +\notag \\&a^2 (e_{yo,t,k}^2 - e_{yo,t,k}\bar{e}_{y,t,k}) - a^2b^2,
    \end{align}
\end{subequations}

\noindent where the ellipse is centered at the front edge of the follower in the desired group and the rear edge of the leader in the current group. As illustrated in Fig.~\ref{change}, the ego vehicle formulates individual ellipse constraints for both vehicles during the LC state and only for the follower during the LP state. These constraints are jointly incorporated into the optimization to ensure safe interactions.

\begin{figure}[t]
\centerline{\includegraphics[width=0.47\textwidth,height=0.46\textwidth]{lcstate.pdf}}
\caption{Constraint mechanisms under different constraint states.} In the LK state, the optimization problem includes longitudinal DCBF constraints for the current lane and lateral DCBF constraints for adjacent lanes. In the LP state, the leading vehicle in the current lane is transformed into a linearized DHOCBF constraint to better accommodate potential lane change maneuvers while actively exploring the feasible space. In the LC state, the rear vehicle in the desired group and the front vehicle in the current group are the main interaction targets. Linearized DHOCBF ensures lane change safety and expands the feasible region for lane changes. 
\label{change}
\end{figure}

Inspired by generalized control barrier function (GCBF) \cite{ma2021feasibility}, which broadens the feasible region for constrained receding horizon control (RHC) by implementing a one-step constraint, we expand the feasible region by applying constraints over a limited number of steps $N_{dho}$, which is less than the total horizon $N$, thereby optimizing feasibility while enhancing the computation efficiency.

Based on \eqref{dhocbfexplan} and \eqref{m2dhocbf}, $\psi_0(\mathbf{x}_{t,k})$ for $k \in \{0,1,..., N_{dho}-1\}$ and $\psi_1(\mathbf{x}_{t,k})$ for $k \in \{0,1,..., N_{dho}-2\}$ are formulated at each step.

\noindent\textbf{\textit{Safety boundary Constraints}}. At no point can the ego vehicle breach the outermost lane boundaries. We constrain the lateral position $e_y$ to remain within the lane boundaries. In addition, when lane keeping, we apply additional constraints to maintain a smaller deviation of the vehicle from the road centerline as follows:
\begin{equation}
    - \lambda_b - \epsilon_b \leq e_y \leq  \lambda_b + \epsilon_b
\end{equation}
where $\lambda_b$ is a small positive constant and $\epsilon_b$ is the slack variable for safety boundary constraints. Safety boundary constraints restrict the lateral error in a certain range centered on the centerline. If the ego vehicle goes far from the centerline,  its cost increases due to a penalty on $\epsilon_b$. In short, we describe this constraint as follows:
\vspace{-0.1cm}
\begin{equation}
    H_{S}(\mathbf{x}) \leq \epsilon_b
\end{equation}
\vspace{-0.1cm}
\noindent\textbf{\textit{Actuator Constraints}}.
Actuator constraints are applied to control inputs to restrict actuator limits:

\begin{equation}
\left\{
\begin{aligned}
a_{x},_{min} \leq a_x \leq a_{x},_{max} \\
\dot{a}_{x},_{min} \leq \dot{a}_x \\ 
\delta_{min} \leq \delta \leq \delta_{max}\\
\dot{\delta}_{min} \leq \dot{\delta} \leq \dot{\delta}_{max}\\ 
\end{aligned}
\right.
\vspace{0.2cm}
\end{equation}

For simplicity, we express actuator constraints as $\mathbf{u} \in \mathcal{U}$, $\dot{\mathbf{u}} \in \mathcal{U}_d$ in the optimal problem formulation.

\subsection{Motion Planner Formulations}
We employed {Frenet-based MPC \cite{huang2023trajectory}} to establish a finite-time convex optimization problem. Based on three different constraint states, we offer three optimization formulations. For simplicity, we abbreviate the three constraint states as LK (Lane-keeping), LP (Lane-probing), and LC (Lane-changing) optimizations. Fig.~\ref{change} provides a concise visual representation of the constraints associated with each of these states, illustrating how the optimization problem adapts to different driving conditions.

\noindent\textbf{\textit{LK Optimization.}} For the lane-keeping state, the optimization problem is formulated as follows:
\begin{subequations}
    \begin{align}
&\underset{\mathbf{u}_t:t+N-1|t,\mathbf{x}_t:t+N|t}{\arg\min}\sum_{k=0}^{N-1} ||\mathbf{u}_{t+k}||_Q + ||\dot{\mathbf{u}}_{t+k}||_P  \notag \\
&+||\mathbf{x}_{t+k}-\mathbf{x}_{ref}||_R + p(\mathbf{x}_{t+N|t})_S + ||\epsilon_{lk}||_{R_{\epsilon}} \\
\text{s.t. }  &\mathbf{x}_{t+k+1|t} = \mathbf{A} \mathbf{x}_{t+k|t} + \mathbf{B} \mathbf{u}_{t+k|t} + \mathbf{C},\label{34b}\\
&\Delta h_{lon}^{cl}(\mathbf{x}_{t+k|t},\mathbf{u}_{t+k|t}) + \gamma_{lon}^{cl} h_{lon}^{cl}(\mathbf{x}_{t+k|t}) \geq \epsilon_{lon}^{cl}, \\
\label{34c}
&\Delta h_{lon}^{cf}(\mathbf{x}_{t+k|t},\mathbf{u}_{t+k|t}) + \gamma_{lon}^{cf} h_{lon}^{cf}(\mathbf{x}_{t+k|t}) \geq \epsilon_{lon}^{cf}, \\
\label{34e}
&\Delta h_{lat}^{\mathcal{J}}(\mathbf{x}_{t+k|t},\mathbf{u}_{t+k|t}) + \gamma_{lat}^{ \mathcal{J}} h_{lat}^{\mathcal{J}}(\mathbf{x}_{t+k|t}) \geq \epsilon_{lat}^{\mathcal{J}}, \\
\label{34f}
&H_{S}(\mathbf{x}_{k+1}) \leq \epsilon_b, \ k=0,...,N-1 \\
&\mathbf{x}_{t|t} = \mathbf{x}_{t}, \\
&\mathbf{x}_{t+k+1|t} \in \mathcal{X}, \ \mathbf{u}_{t+k+1|t} \in \mathcal{U}, \ k=0,...,N-1, \\
&\dot{\mathbf{u}}_{t+k+1|t} \in \mathcal{U}_d, \ k=0,...,N-2 
    \end{align}
\end{subequations}
In the LK state, the optimization constraints include longitudinal constraints for the leader $\Delta h_{lon}^{cl}$ and follower $\Delta h_{lon}^{cf}$ vehicles at the current node, as well as lateral constraints $\Delta h_{lat}^{\mathcal{J}}$ for surrounding vehicles within the ROI.

\noindent\textbf{\textit{LP Optimization.}} In the context of lane-probing, the optimization framework is structured as follows:
\vspace{0.1cm}
\begin{subequations}
    \begin{align}
&\underset{\mathbf{u}_t:t+N-1|t,\mathbf{x}_t:t+N|t}{\arg\min}\sum_{k=0}^{N-1} ||\mathbf{u}_{t+k}||_Q + ||\dot{\mathbf{u}}_{t+k}||_P \notag \\
&+ ||\mathbf{x}_{t+k}-\mathbf{x}_{ref}||_R + p(\mathbf{x}_{t+N_p|t})_S + ||\epsilon_{lp}||_{R_{\epsilon}} \\
\text{s.t. }  &\mathbf{x}_{t+k+1|t} = \mathbf{A} \mathbf{x}_{t+k|t} + \mathbf{B} \mathbf{u}_{t+k|t} + \mathbf{C},\label{35b}\\ 
&\Delta h_{lon}^{cf}(\mathbf{x}_{t+k|t},\mathbf{u}_{t+k|t}) + \gamma_{lon}^{cf} h_{lon}^{cf}(\mathbf{x}_{t+k|t}) \geq \epsilon_{lon}^{cf}, \\
&\Delta h_{lat}^{\mathcal{J}}(\mathbf{x}_{t+k|t},\mathbf{u}_{t+k|t}) + \gamma_{lat}^{\mathcal{J}} h_{lat}^{\mathcal{J}}(\mathbf{x}_{t+k|t}) \geq \epsilon_{lat}^{\mathcal{J}}\label{35e}\\
&\psi_{0}^{cl}(x_{t,k+1}) \geq w_{t,k+1,1}(1-\gamma_1^{cl})^{k+1}\psi_{0}^{cl}(x_{t,0}),\\
&\psi_{1}^{cl}(x_{t,k+1}) \geq (1-\gamma_2^{cl})^{k+1}\psi_{0}^{cl}(x_{t,1}) \\
&\quad + w_{t,k+1,2}(\gamma_1-1)(1-\gamma_2^{cl})^{k+1}\psi_{0}^{cl}(x_{t,0}),\\
&H_{S}(\mathbf{x}_{k+1}) \leq \epsilon_b, \ k=0,...,N-1\\
&\mathbf{x}_{t|t} = \mathbf{x}_{t}\\
&\mathbf{x}_{t+k+1|t} \in \mathcal{X}, \ \mathbf{u}_{t+k+1|t} \in \mathcal{U}, \ k=0,...,N-1 \\
&\dot{\mathbf{u}}_{t+k+1|t} \in \mathcal{U}_d, \ k=0,...,N-2 
    \end{align}
\end{subequations}
In the LP state, the optimization constraints include longitudinal constraints for the follower $\Delta h_{lon}^{cf}$ at the current node, lateral constraints $\Delta h_{lat}^{\mathcal{J}}$ for surrounding vehicles within the ROI. Different from the LK state, the constraints for the leader vehicle at the current node are ellipse constraints, $\psi_{0}^{cl}$ and $\psi_{1}^{cl}$.

\noindent\textbf{\textit{LC Optimization.}} Within the lane-changing scenario, the optimization framework is organized as follows:
\vspace{-0.05cm}
\begin{subequations}
    \begin{align}
&\underset{\mathbf{u}_t:t+N-1|t,\mathbf{x}_t:t+N|t}{\arg\min}\sum_{k=0}^{N-1} ||\mathbf{u}_{t+k}||_Q + ||\dot{\mathbf{u}}_{t+k}||_P \notag \\
\label{36a}
&+||\mathbf{x}_{t+k}-\mathbf{x}_{ref}||_R + p(\mathbf{x}_{t+N|t})_S + ||\epsilon_{lc}||_{R_{\epsilon}} \\
\text{s.t. }  &\mathbf{x}_{t+k+1|t} = \mathbf{A} \mathbf{x}_{t+k|t} + \mathbf{B} \mathbf{u}_{t+k|t} + \mathbf{C}, \label{36b}\\
\label{36c}
&\Delta h_{lon}^{cf}(\mathbf{x}_{t+k|t},\mathbf{u}_{t+k|t}) + \gamma_{lon}^{cf} h_{lon}^{cf}(\mathbf{x}_{t+k|t}) \geq \epsilon_{lon}^{cf}, \\
\label{36d}
&\Delta h_{lon}^{dl}(\mathbf{x}_{t+k|t},\mathbf{u}_{t+k|t}) + \gamma_{lon}^{dl} h_{lon}^{dl}(\mathbf{x}_{t+k|t}) \geq \epsilon_{lon}^{dl},\\
\label{36e}
&\Delta h_{lat}^{\mathcal{J}}(\mathbf{x}_{t+k|t},\mathbf{u}_{t+k|t}) + \gamma_{lat}^{\mathcal{J}} h_{lat}^{\mathcal{J}}(\mathbf{x}_{t+k|t}) \geq \epsilon_{lat}^{\mathcal{J}}\\
\label{36f}
&\psi_{0}^{cl}(x_{t,k+1}) \geq w_{t,k+1,1}(1-\gamma_1^{cl})^{k+1}\psi_{0}^{cl}(x_{t,0}),\\
\label{36g}
&\psi_{0}^{df}(x_{t,k+1}) \geq w_{t,k,1}(1-\gamma_1^{df})^{k+1}\psi_{0}^{df}(x_{t,0})\\
\label{36h}
&\psi_{1}^{cl}(x_{t,k+1}) \geq (1-\gamma_2^{cl})^{k+1}\psi_{0}^{cl}(x_{t,1})\notag \\
&\quad + w_{t,k+1,2}(\gamma_1^{cl}-1)(1-\gamma_2^{cl})^{k+1}\psi_{0}^{cl}(x_{t,0}),\\
\label{36i}
&\psi_{1}^{df}(x_{t,k+1}) \geq (1-\gamma_2^{df})^{k+1}\psi_{0}^{df}(x_{t,1}) \notag\\
&\quad + w_{t,k+1,2}(\gamma_1^{df}-1)(1-\gamma_2^{df})^{k+1}\psi_{0}^{df}(x_{t,0})\\
\label{36j}
&H_{S}(\mathbf{x}_{k+1}) \leq \epsilon_b,  \ k=0,...,N-1\\
\label{36k}
&\mathbf{x}_{t|t} = \mathbf{x}_{t}\\
\label{36l}
&\mathbf{x}_{t+k+1|t} \in \mathcal{X}, \mathbf{u}_{t+k+1|t} \in \mathcal{U}, \ k=0,...,N-1\\
\label{36m}
&\dot{\mathbf{u}}_{t+k+1|t} \in \mathcal{U}_d, \ k=0,...,N-2
    \end{align}
\end{subequations}

In the LC state, the optimization constraints include longitudinal constraints for the desired leader $\Delta h_{lon}^{dl}$ and current folower $\Delta h_{lon}^{cf}$, as well as ellipse constraints for both the desired follower $\psi_{0}^{df},\psi_{1}^{df}$ and the current leader $\psi_{0}^{cl},\psi_{1}^{cl}$, represented by the corresponding ellipse functions. Additionally, lateral constraints $\Delta h_{lat}^{\mathcal{J}}$ are applied to surrounding vehicles within the ROI to ensure safe lateral separation.

Among this three optimization formulation, (\ref{34b}), (\ref{35b}), (\ref{36b}) denote the constraints imposed by system dynamics.

In the objectives in these three optimization formulation, the vector $\mathbf{x}_{ref}$ represents the desired values for $v_x$, $w$, $v_y$, along with the desired values for $e_y$ and $e_{\psi}$, which are set to $0$. This setting arises because, in Frenet coordinates, all points are projected onto the centerline that the ego vehicle is intended to follow, thereby making the desired deviation distance and heading angle error equal to zero. $p(\mathbf{x}_{t+N_p|t})$ is the terminal cost, where $S$ signifies the weight attributed to the terminal cost. Drawing upon the method described by \cite{zeng2021safety} that employs a control Lyapunov function (CLF) as the terminal cost, we utilize the squared error $e_{\psi}^2$ as the terminal cost. $||\mathbf{u}_{t+k}||$, $||\dot{\mathbf{u}}_{t+k}||$, and $||\mathbf{x}_{t+k} - \mathbf{x}_{ref}||$ serve as stage costs, with $Q$, $P$, and $R$ representing the weights for the control inputs, input derivatives, and quadratic terms of tracking $\mathbf{x}_{\text{ref}}$, respectively. In particular, the weight matrix \( R \) for \( v_x \), \( e_{\psi} \), and \( e_y \) is designed to increase over time. This time-increasing strategy helps to prevent the generation of excessively large steering angles at the beginning of the trajectory.

\section{Experiment}
In this section, we perform comprehensive simulations to evaluate the proposed system IDEAM. In section \ref{sec:A}, we introduce the simulation background about scenarios and parameters. In section \ref{sec:B}, we present the simulation benchmark including traditional methods and speed-oriented approaches, and evaluation metrics. Section \ref{sec:C} gives the simulation results, and section \ref{sec:D} shows a detailed case study of simulations.
\label{sec:experinment}
\subsection{Scenario and Parameters Description}
\label{sec:A}

 \begin{table}[t]
\centering
\captionsetup{
  justification=centering, 
  labelsep=space, 
  textfont=sc, 
  labelfont=sc, 
  format=plain 
}
\caption{Parameters of Vehicle Model}
\label{tab:vehicle-parameters}
\renewcommand{\arraystretch}{1.2} 
\begin{tabular}{@{}llc@{}}
\toprule
\toprule
Symbol & Description & Value \\ \midrule
$m$ & Vehicle mass & 1292 \si{kg} \\
$I_z$ & Vehicle yaw moment of inertia & 1343.1 \si{kg\cdot m^2} \\
$l_f$ & Distances from the front axles to the CoG & 1.56 \si{m} \\
$l_r$ & Distances from the rear axles to the CoG & 1.04 \si{m} \\
$l$ & Length of ego vehicle & 3.5 \si{m} \\
$a_{a,max}$ & Maximum acceleration & 3.00 \si{m/s^2} \\
$a_{d,max}$ & Maximum deceleration & -3.00 \si{m/s^2} \\
$\delta_{max}$ & Maximum steering angle & 0.44 \si{rad} \\
$v_{d}$ & Desired velocity of ego vehicle & 18.00 \si{m/s} \\
\bottomrule
\bottomrule
\end{tabular}
\end{table}


\noindent\textbf{\textit{Simulation Scenarios.}} The simulation scenario occurs on a continuous closed-loop path that includes both curves and straight sections, challenging the autonomous system to maintain consistent performance across the varying road geometry. The surrounding vehicles are randomly generated within the scene, with varying initial positions and desired velocities, ensuring a diverse set of interactions. The longitudinal motion of surrounding vehicles is simulated using the intelligent driver model (IDM) \cite{treiber2013traffic}, while lateral control is achieved through proportional-integral-derivative (PID) controller, the entire scenario takes place within a multi-lane environment, where the curvature of the bends varies along the path, reaching a maximum of approximately $0.1$. The dynamics of the scenario are updated at each step with a discretization interval of $\Delta t = 100$ $\mathrm{ms}$, following a kinematic model that dictates vehicle movements as outlined below,
\vspace{0.2cm}
\begin{subequations}
\begin{align}
\dot{x} &= v \cos(\psi + \beta) \label{1a} \\
\dot{y} &= v \sin(\psi + \beta) \label{1b} \\
\dot{\psi} &= \frac{v}{l_r} \sin\beta \label{1c} \\
\dot{v} &= a \label{1d} \\
\beta &= \tan^{-1}\left(\frac{l_r}{l_f+l_r} \tan \delta_f\right) \label{1e}
\end{align}
\end{subequations}

To thoroughly evaluate the system's performance, we conducted 200 parallel simulations under these conditions.

\noindent\textbf{\textit{Parameters.}} Table \ref{tab:vehicle-parameters} contains the parameters of the vehicle's model in this paper. The parameters of the LSGM algorithm and motion planner are detailed in Table \ref{tab:algorithm parameters}.


\subsection{Benchmark and Metrics}

\label{sec:B}
\noindent\textbf{\textit{Baselines.}} To validate the effectiveness and advantages of the proposed IDEAM framework, we compare it with the following baseline methods using $200$ parallel tracks for each method, resulting in a total of $1200$ simulations:

\begin{enumerate}
    \item \textit{SO-DM} \cite{shu2023safety} The speed-oriented decision-making model guides autonomous vehicles in lane-changing to strive for speed benefits while ensuring safety. We will compare the performance with $LAS=60$ (the parameter that has shown good performance) and $LAS=30$, as a lower $LAS$ value may perform better in dense traffic conditions.
    \item \textit{DRB-FSM} \cite{chen2020hierarchical} FSM decision-making for DBR in \cite{chen2020hierarchical} selects appropriate driving behaviors through state transitions and coordinates vehicle motion control, enhancing driving adaptability.
    \item \textit{No-Probing IDEAM} The No-Probing IDEAM system refers to our proposed IDEAM framework implemented without active probing in the motion planner. By comparing the No-Probing IDEAM system with the full IDEAM framework, we aim to highlight the speed advantages gained through the inclusion of probing mechanisms in the motion planner.
    \item \textit{MOBIL} \cite{kesting2007general} MOBIL is a well-known traditional lane change strategy designed to improve overall traffic flow efficiency, the speed benefits for the ego vehicle, and enhance safety. It achieves these goals by carefully managing lane changes to minimize the braking impact on surrounding vehicles, thereby maintaining steady traffic conditions.  This method is primarily focused on optimizing traffic flow in typical driving scenarios, distinguishing it from methods designed specifically for high-priority situations like emergency autonomous vehicles.
\end{enumerate}
\begin{table}[t]
\centering
\captionsetup{
  justification=centering, 
  labelsep=space, 
  textfont=sc, 
  labelfont=sc, 
  format=plain 
}
\caption{Parameters of LSGM algorithm and Motion Planner}
\label{tab:algorithm parameters}
\renewcommand{\arraystretch}{1.35}
\setlength{\tabcolsep}{4pt}
\begin{tabular}{c|c|c|c}
\toprule
\toprule
$Q$ & $\text{diag}(0.1,8.0)$ & $S$ & $60.0$ \\
$P$ & $\text{diag}(0.0,8.0) $ & $T_{\text{long}}$ & $70$ \\
$R$ & $\text{diag}([0.2,0.35],4.0,4.0,0.0,[6,10],[7,24])$ & $T_{\text{short}}$ & $[20,60]$ \\
$N$ & $30$ & $d_{lat}$ & $2.1$ \si{m} \\
$N_{dho}$ & $20$ & $\epsilon$ & $3.5$ \si{m} \\
$a_l,a_f$ & $1.5,1.5$ & $n$ & $3$ \\
$b_l,b_f$ & $2.2,2.3 $ & $T_d$ & $0.3$ \\
$l_{diag}$ & $3.7$ \si{m}  & $d_0$ & $5$ \si{m}\\
\bottomrule
\bottomrule
\end{tabular}
\end{table}

\sisetup{detect-weight=true, detect-family=true, mode=text}

\begin{table*}[t]
\centering
\captionsetup{
  justification=centering, 
  labelsep=space, 
  textfont=sc, 
  labelfont=sc, 
  format=plain 
}
\caption{Performance Comparison between six systems Boxplot of 40s Progress}
\label{tab:result}
\setlength{\tabcolsep}{4pt}
\renewcommand{\arraystretch}{1.25}
\begin{tabular}{
  @{}l
  S[table-format=3.2]
  S[table-format=3.2]
  S[table-format=3.2]| 
  S[table-format=2.2]
  S[table-format=2.2]| 
  S[table-format=1.2]
  S[table-format=1.2]| 
  S[table-format=1.2]
  S[table-format=1.2]
  S[table-format=1.2]
@{}}
\toprule
\toprule
\multirow{2}{*}{\parbox{1cm}{\centering Method}} & 
\multicolumn{1}{c}{$20s$ Prog.} & 
\multicolumn{1}{c}{$40s$ Prog.} & 
\multicolumn{1}{c|}{Max. Prog.} & 
\multicolumn{1}{c}{Avg. Vel.} & 
\multicolumn{1}{c|}{Max. Vel.} & 
\multicolumn{1}{c}{Avg Min $\mathcal{S}_{o}$} & 
\multicolumn{1}{c|}{Min. $\mathcal{S}_{o}$} & 
\multicolumn{1}{c}{Max. Acc.} & 
\multicolumn{1}{c}{Avg. Acc.} &
\multicolumn{1}{c}{Avg. Jerk.} \\ 
& {$(\mathrm{m})$} & {$(\mathrm{m})$} & { $(\mathrm{m})$} & {$(\mathrm{m/s})$} & {$(\mathrm{m/s})$} & {$(\mathrm{m})$}  & {$(\mathrm{m})$} & { $(\mathrm{m/s}^2)$} & { $(\mathrm{m/s}^2)$} &{ $(\mathrm{m/s}^3)$} \\
\midrule
SO-DM ($LAS=30$)  & 251.60 & 467.84 & 578.13 & 11.38 & 18.12 & 2.58 & 1.64 & 3.00 & 0.95 & 0.12\\
SO-DM ($LAS=60$) & 250.82 & 466.61 & 578.13 & 11.34 & 18.37 & \textbf{2.65} & 1.71 & 3.00 & \textbf{0.87}& 0.09\\
DRB-FSM & 250.83 & 465.06 & 548.21 & 11.33 & 18.12 & \textbf{2.65} & 1.68 & 3.00 & \textbf{0.87}& \textbf{0.08}\\
MOBIL & 253.61 & 475.03 & 615.53 & 11.52 & 19.85 & 2.50 & \textbf{1.75} & 3.00 & 0.97 & 0.11\\
No-Probing IDEAM & 257.26 & 498.97 & 613.52 & 11.91 & 20.74 & 2.33 & 0.84 & 3.00 & 1.27 & 0.13\\
IDEAM (ours) & \textbf{265.08} & \textbf{525.19} & \textbf{619.60} & \textbf{12.44} & \textbf{21.32} & 2.26 & 1.14 & 3.00 & 1.40 & 0.22\\
\bottomrule
\bottomrule
\end{tabular}
\end{table*}

All simulations are conducted on a laptop with a $2.70$ GHz Intel Core i7-12700H processor and $16$ GB RAM. We use cvxpy \cite{diamond2016cvxpy} and ECOS \cite{domahidi2013ecos} solver with Python $3.7$ to solve the QP problem in MPC. Additionally, SO-DM, DRB-FSM, and MOBIL utilize the planner of No-probing IDEAM without the time-increasing strategy, ensuring fairness in the comparison.

\noindent\textbf{\textit{Metrics.}} We evaluate these methods by comparing metrics related to progress, efficiency, safety, and comfort across $200$ parallel tracks. This comprehensive assessment allows us to statistically analyze the performance of each method under various conditions, providing a detailed comparison of their effectiveness:
\begin{enumerate}
\item \textit{Progress:} Progress serves as a key indicator of how well the vehicle is advancing along its intended path, offering insights into its ability to navigate different scenarios effectively. By evaluating progress, we gain a clearer understanding of the vehicle's movement relative to its goal. To evaluate progress, we project the vehicle's center onto the lane centerline and measure the average longitudinal distances across all tracks at $20$ and $40$ seconds. These measurements reflect the average progress across all tracks at those moments. We also assess the maximum progress achieved within the $200$ tracks to provide a comprehensive view of performance.
\item \textit{Efficiency:} Efficiency is particularly crucial for emergency vehicles, where every second counts. It reflects the vehicle's ability to operate swiftly and without delay.  Therefore, measuring efficiency provides critical insights into how well the vehicle optimizes its movements under various conditions. We measure efficiency by calculating both the average and maximum longitudinal velocity of the ego vehicle at every step across all tracks throughout the entire process.
\item \textit{Safety:} Maintaining safety is essential in ensuring that the vehicle can navigate high-interaction environments without incident. Effective safety management requires careful monitoring of vehicle spacing to avoid potential collisions. In this paper, safety will be assessed by calculating the minimum distance $S_o$ between the ego vehicle and all surrounding vehicles at every step across all tracks, as well as the average of these minimum distances.
\item \textit{Comfort:} To measure comfort, we assess the vehicle's dynamics by evaluating the maximum acceleration, the average acceleration, and the average jerk across all tracks. These metrics provide insights into the smoothness and stability of the ride.
\end{enumerate}

\subsection{Statistical Results}
\label{sec:C}
We compare our methods against the baselines discussed in Section~\ref{sec:B}. The statistical results for the evaluated metrics are summarized in Table~\ref{tab:result}.

\textbf{Progress:} In terms of progress, the IDEAM performs best among all evaluation metrics. In the average $20~\mathrm{s}$ and $40~\mathrm{s}$ progress measurements, IDEAM achieved $265.08$ meters and $525.19$ meters, respectively, significantly ahead of other methods, indicating its continued superiority on long-time scales. Compared with No-probing IDEAM, IDEAM demonstrated a clear advantage, with a difference of $7.82$ meters at $20~\mathrm{s}$ and $26.22$ meters at $40~\mathrm{s}$. This difference highlights the benefits of incorporating lane probing in the motion planner design, allowing IDEAM to make more informed decisions and optimize vehicle trajectory more effectively. In comparison to SO-DM, DRB-FSM, and MOBIL, IDEAM opened up an even larger gap. At $20~\mathrm{s}$, IDEAM extended its lead over these methods by a range of approximately $11$ to $14$ meters. By $40~\mathrm{s}$, this gap widened further, with IDEAM ahead by about $50$ to $60$ meters. These results highlight the effectiveness of the IDEAM method in maintaining superior progress over both short and long time scales compared to the other baseline methods.  Furthermore, the maximum progress of the IDEAM is $619.60$ meters, which is also higher than all baselines.

\textbf{Efficiency:} Regarding efficiency, IDEAM reached an average velocity of $12.44$ $~\mathrm{m/s}$, outperforming all other methods. It maintained a speed advantage of approximately $0.53~\mathrm{m/s}$ to $1.11~\mathrm{m/s}$ over the others. Additionally, IDEAM achieved the highest maximum velocity of $21.32~\mathrm{m/s}$, further demonstrating its capability to optimize vehicle performance across different conditions. This superior speed and acceleration highlight IDEAM's strong potential in emergency scenarios, where minimizing response time and enhancing operational efficiency are crucial.

\textbf{Safety:} Safety was assessed through the minimum safety distance and the average minimum safety distance. Table~\ref{tab:result} indicates that the SO-DM and DRB-FSM methods performed best in maintaining an average minimum safety distance of $2.65$ meters, showing strong capability in keeping a safe distance from surrounding objects. Although IDEAM had a slightly lower average minimum safety distance of $2.26$ meters, it still preserved a reasonable safety margin. When looking at the minimum safety distance, the MOBIL method showed the best performance with $1.75$ meters, while IDEAM recorded $1.14$ meters, the values of IDEAM still fall within safe limits, ensuring that the system maintains adequate safety.

\textbf{Comfort:} Comfort is evaluated by analyzing vehicle dynamics, including maximum acceleration, average acceleration, and average jerk. The table shows that all methods have the same maximum acceleration of $3.00~\mathrm{m/s}^2$. In terms of average acceleration, the SO-DM ($LAS=60$) and DRB-FSM methods performed best, both with the lowest average acceleration of $0.87~\mathrm{m/s}^2$. In contrast, IDEAM had a higher average acceleration of $1.40~\mathrm{m/s}^2$. Regarding average jerk, which measures the smoothness of acceleration, the DRB-FSM method performed the best with a value of $0.08~\mathrm{m/s}^3$, offering the smoothest driving experience. IDEAM had a higher average jerk of $0.22$ $~\mathrm{m/s}^3$, which indicates a slight trade-off in comfort compared to SO-DM, DRB-FSM, and MOBIL. These results suggest that while IDEAM maintains strong performance in efficiency and progress, it shows a modest difference in ride smoothness.

\begin{figure*}[t]
    \centering
    \includegraphics[width=\textwidth]{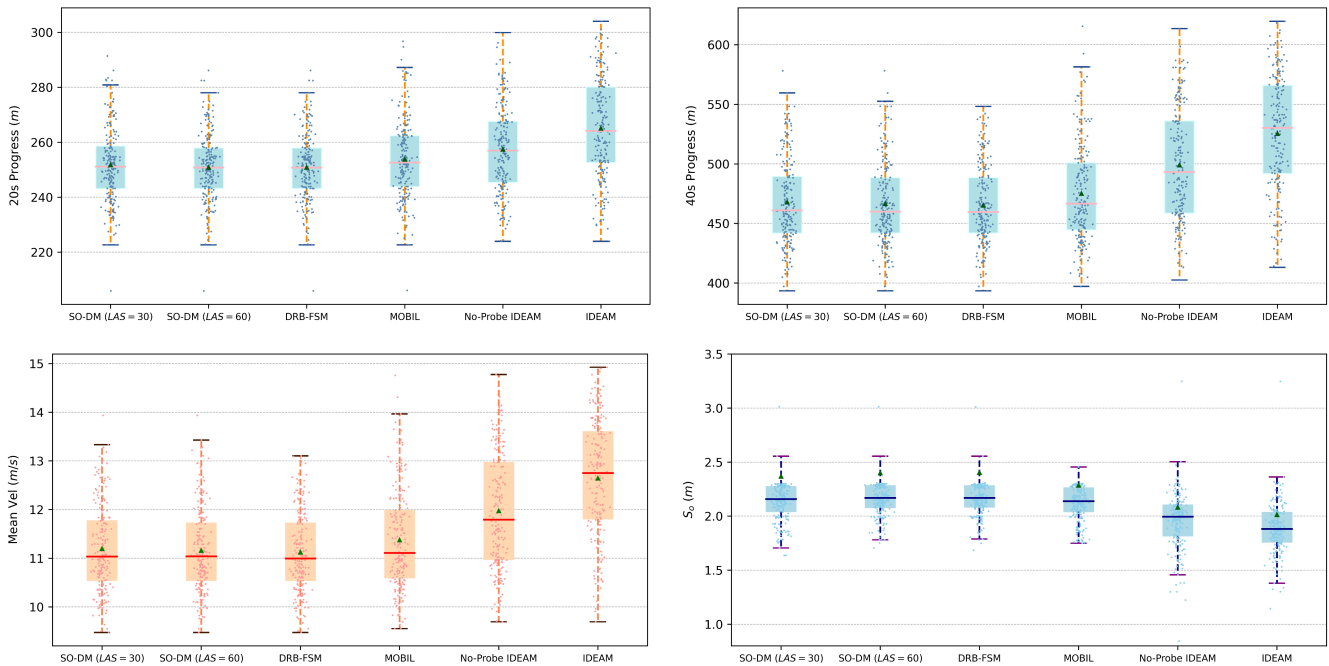} 
     \caption{Boxplots showing the $20s$ and $40s$ progress, velocity, and $S_o$ comparisons between IDEAM and baseline methods.} 
     \label{fig:box}  
\end{figure*}

 Fig.~\ref{fig:box} shows the boxplots of the $20~\mathrm{s}$ and $40~\mathrm{s}$ progress, velocity, and $S_o$ (safety) metrics of the proposed IDEAM method compared against other baseline methods. Although IDEAM exhibits slightly smaller $S_o$ values, these values remain within acceptable safety margins, ensuring that the system maintains a sufficient level of safety while delivering significant improvements in efficiency and progress.

\textbf{Lane Change Intention:} Table~\ref{tab:Lane Change Intention} shows that IDEAM achieved the highest average lane change success of $5$ ($4.74$) times, with a maximum success of $11$ times, exceeding all other methods. The higher frequency of lane changes suggests that IDEAM is more responsive to traffic conditions, likely indicating an active strategy to explore better positioning on the road. This tendency to change lanes more often may reflect an effort to create opportunities for increased speed and efficiency, particularly in complex traffic scenarios.

\begin{table}[h]
\centering
\captionsetup{
  justification=centering, 
  labelsep=space, 
  textfont=sc, 
  labelfont=sc, 
  format=plain 
}
\caption{Lane Change Intention}
\label{tab:Lane Change Intention}
\setlength{\tabcolsep}{6pt} 
\renewcommand{\arraystretch}{1.2} 
\begin{tabular}{@{}lcccc@{}}
\toprule
\toprule
Method  & Avg. LC. success & Max. LC. success \\ \midrule
SO-DM (LAS=30) & 1 (1.11) & 3 & \\
SO-DM (LAS=60) & 1 (1.00)& 4 & \\
DRB-FSM & 1 (0.95) & 4 & \\
MOBIL & 2 (1.83) & 6 & \\
No Probing-IDEAM & 4 (3.75) & 9 & \\
IDEAM (ours) & \textbf{5 (4.74)} & \textbf{11} & \\
\bottomrule
\bottomrule
\end{tabular}
\end{table}

\textbf{Computation Time:} As shown in Fig.~\ref{fig:time}, the LSGM algorithm achieves an average execution time of $0.34~\mathrm{ms}$ and a maximum of $2.01~\mathrm{ms}$, which includes the runtime of the DFS module. These values are based on statistics collected from each frame across several scenarios. The DFS module itself incurs negligible computational cost, with an average close to zero and a maximum of $1.17~\mathrm{ms}$. The solver exhibits an average execution time of $5.97~\mathrm{ms}$ and a maximum of $21.03~\mathrm{ms}$.

\begin{figure}[h]
    \includegraphics[width=0.5\textwidth]{time.pdf} 
    \caption{Computation Time of the solver, LSGM algorithm, and the C-DFS algorithm.}  
     \label{fig:time}  
\end{figure}

\subsection{Case Study}
\label{sec:D}
In this section, two types of scenarios, namely evaluation and emergency scenarios, are presented.

\begin{figure*}[t]
    \centering
    \includegraphics[width=\textwidth]{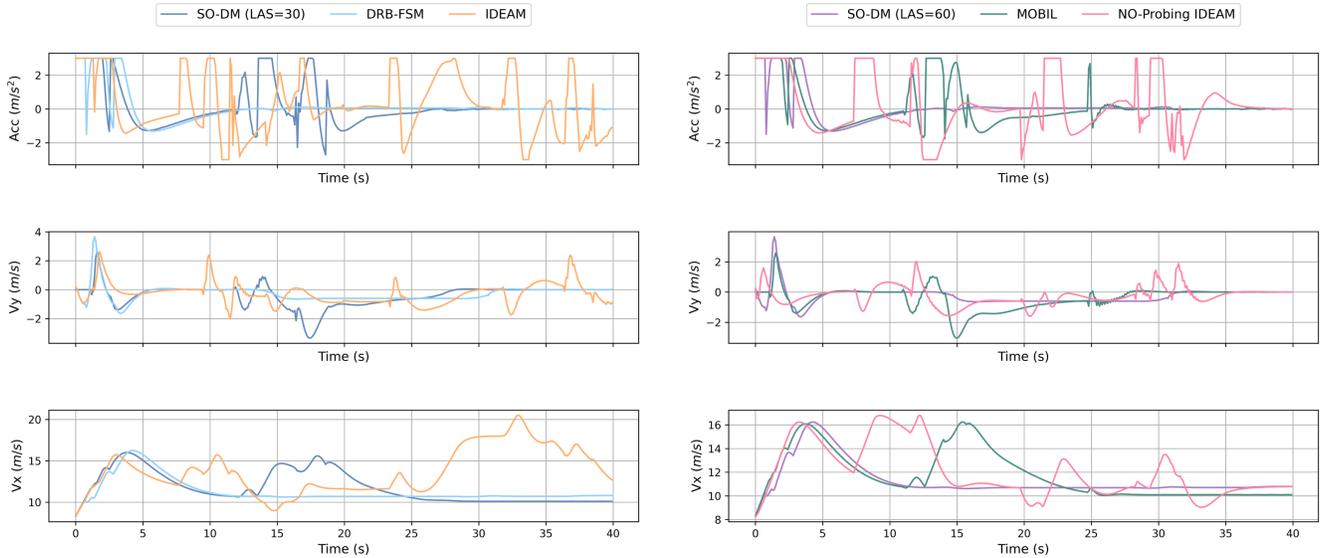} 
    \caption{Comparison of acceleration, lateral velocity, and longitudinal velocity of IDEAM and baseline methods. IDEAM generally maintains higher longitudinal velocity compared to the baselines, while all methods exhibit frequent acceleration changes to balance efficiency and safety.}  
     \label{fig:case acc}  
\end{figure*}

\begin{figure*}[htbp]
    \centering
    \includegraphics[width=\textwidth]{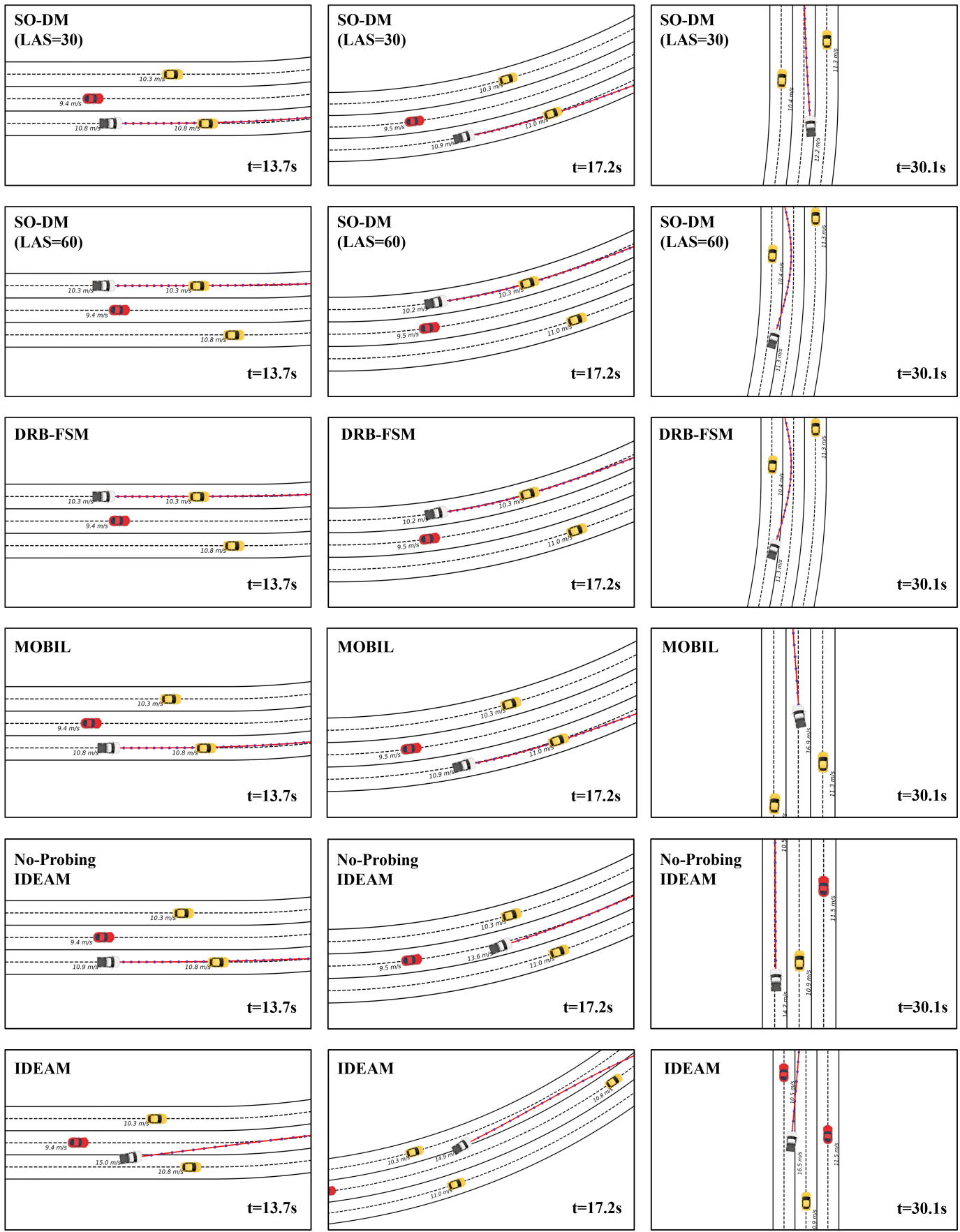} 
    \caption{Snapshots from simulations comparing IDEAM system with other baseline methods at $t=13.7s$, $t=17.2s$, $t=30.1s$.}   
    \label{fig:case study}  
\end{figure*}

\noindent\textbf{Evaluation Scenario} We randomly select a scenario to intuitively demonstrate the effectiveness of the proposed method. Specifically, we selected three key frames at $t=13.7~\mathrm{s}$, $t=17.2~\mathrm{s}$, and $t=30.1~\mathrm{s}$, as shown in Fig.~\ref{fig:case study}. 
At $t=13.7~\mathrm{s}$, after lane-probing, IDEAM guided the ego vehicle to make an early left lane change for speed gain. In contrast, No-Probing IDEAM and MOBIL delayed its lane change due to the lack of spatial exploration. SO-DM ($LAS=60$) and DRB-FSM also performed early left lane changes, benefiting from a faster leading vehicle ($11.3~\mathrm{m/s}$), whereas SO-DM ($LAS=30$) prioritized spatial gaps and shifted to the right lane later.
By $t=17.2~\mathrm{s}$, IDEAM had already completed another left lane change, while No-Probing IDEAM had just reached the middle lane, and other baselines remained unchanged due to conservative or passive decisions. At $t=30.1~\mathrm{s}$, IDEAM completed another lane change, overtaking the yellow vehicle in the middle lane and achieving the greatest progress. Conversely, MOBIL and other baselines lagged without effectively completing the overtaking maneuver. Notably, SO-DM ($LAS=30$) outperformed SO-DM ($LAS=60$) by making better use of spatial benefits.
Overall, IDEAM completed five lane changes, while SO-DM ($LAS=30$), SO-DM ($LAS=60$), DRB-FSM, and MOBIL performed three, two, two, and three lane changes, respectively. Although No-Probing IDEAM also executed five lane changes, its performance was less efficient, demonstrating the advantage of proactive lane-probing. All methods' acceleration, lateral velocity, and longitudinal velocity profiles are shown in Fig.~\ref{fig:case acc}.

\noindent\textbf{Emergency Scenario} As illustrated in Fig.~\ref{fig:emeregncy_two_scenarios}, two emergency scenarios are presented where the ego vehicle encounters sudden deceleration of a leading vehicle. In the first scenario, the leading vehicle gradually slows down, prompting the ego vehicle to reduce its speed while actively exploring more favorable lanes. At $t=13.2~\mathrm{s}$, it performs a left lane change, safely avoids the slowing vehicle, and subsequently overtakes it to regain speed advantages. In the second scenario, the leading vehicle undergoes a sudden and sharp deceleration, forcing the ego vehicle to promptly slow down to avoid a potential collision. Despite the urgency, the ego vehicle continues to search for safer and more efficient lanes and successfully executes a left lane change at $t=16.0~\mathrm{s}$ to bypass the decelerating vehicle. It then overtakes the lead vehicle and restores its driving efficiency.
\section{Discussion}
\label{sec:discussion}

The IDEAM framework has several limitations that need to be addressed. These limitations lead us to consider future work in improving the system's ability. Firstly, it does not fully account for the various sources of uncertainty, such as environmental factors, sensor noise, and interaction uncertainties. Environmental uncertainties, like road conditions, may significantly influence vehicle behavior modeling in motion planning. While methods like Gaussian process regression may help model these uncertainties, improving the vehicle dynamics simulation's accuracy, further work is needed to refine this integration.

Additionally, interaction uncertainty, especially the variability in reaction times, plays a crucial role in decision-making. As highlighted in \cite{wiseman2024autonomous}, human reaction times tend to average around $1.3~\mathrm{s}$, while autonomous vehicles respond in approximately $100$ milliseconds. This difference may be incorporated into the DHOCBF and DCBF formulations to improve real-world adaptability. Furthermore, sensor noise and perception errors affect tracking and prediction, and could potentially be mitigated by integrating techniques such as Kalman filtering into the planner to enhance the planner's performance in uncertain environments. Future work could aim to refine these approaches to create a more robust and adaptive system.

Furthermore, the treatment of interacting agents can be further developed. Game-theoretic approaches could be employed to model and optimize interactions among agents, enhancing the system’s adaptability in high-interaction environments.

\section{Conclusion}
\label{sec:conclusion}
In this paper, we introduced IDEAM, a system designed to enhance both efficiency and safety for emergency autonomous vehicles. The system integrates two key components: the LSGM decision-making algorithm, which prioritizes speed by generating efficient paths, and the motion planner, which adjusts constraints using DCBFs and DHOCBFs within the MPC framework. This motion planner allows the vehicle to actively explore spatial advantages while maintaining safety in dense traffic scenarios. Our experiments demonstrate that IDEAS significantly improves efficiency compared to baseline methods, with some necessary trade-offs in comfort. Future work could explore integrating large language models (LLMs) to assist in selecting specific goals or navigation trajectories for desired groups, potentially reducing jerk and further optimizing both efficiency and driving smoothness. Additionally, incorporating reinforcement learning (RL) in the LSGM algorithm to learn optimal parameters dynamically could further enhance decision-making adaptability and performance in various traffic conditions.








\begin{figure*}[htbp]
    \centering
    \begin{subfigure}[b]{\textwidth}
        \centering
        \includegraphics[width=\textwidth]{emergency_case.pdf}
        \caption{Snapshots from the first emergency scenario.}
        \label{fig:image_a}
    \end{subfigure}
    
    \vspace{1em}  

    \begin{subfigure}[b]{\textwidth}
        \centering
        \includegraphics[width=\textwidth]{emergency_case2.pdf}
        \caption{Snapshots from the second emergency scenario.}
        \label{fig:image_b}
    \end{subfigure}

    \caption{Snapshots from two emergency scenarios at certain timestamps.}
    \label{fig:emeregncy_two_scenarios}
\end{figure*}


\ifCLASSOPTIONcaptionsoff
  \newpage
\fi



%


\vspace{-1cm}
\begin{IEEEbiography}
[{\includegraphics[width=1in,height=1.25in,clip,keepaspectratio]{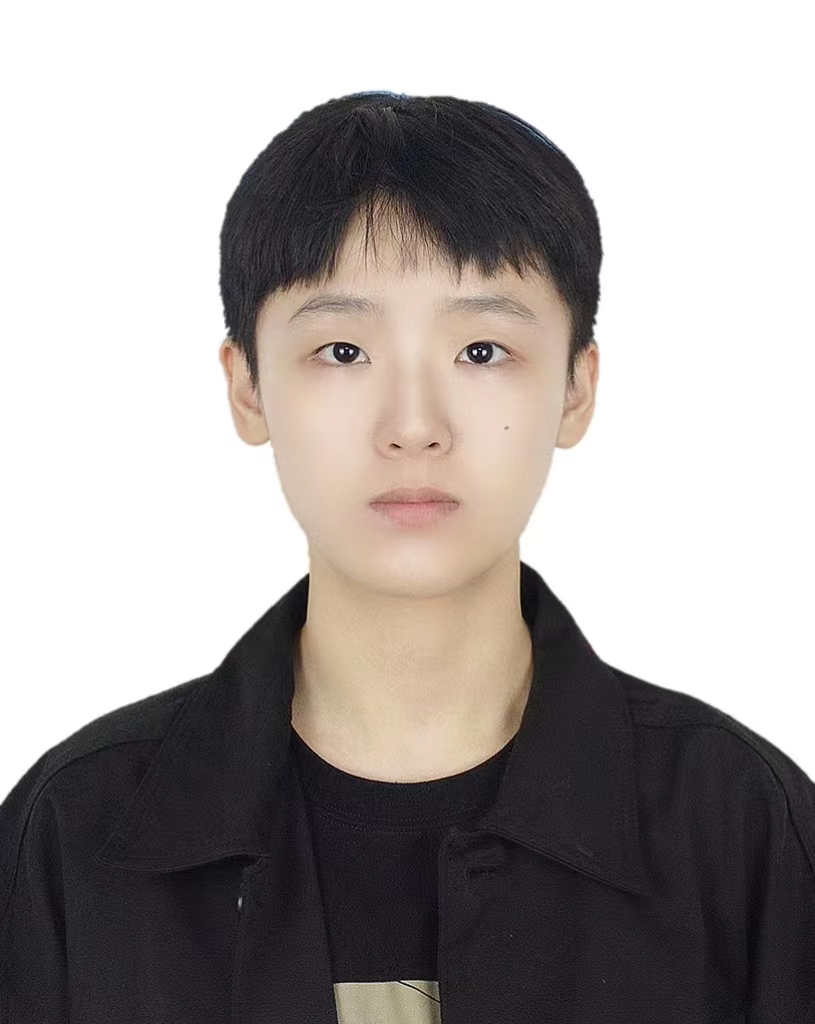}}]{Yiming Shu}
received the B.Eng. degree in Automotive Engineering from the Harbin Institute of Technology (Weihai), Weihai, China, in 2022. She is currently working towards an M.Phil. degree with the University of Hong Kong (HKU). Her research interests include safety-critical motion planning and decision-making of autonomous vehicles (AVs).
\end{IEEEbiography}
\vspace{-1cm}
\begin{IEEEbiography}
[{\includegraphics[width=1in,height=1.25in, clip,keepaspectratio]{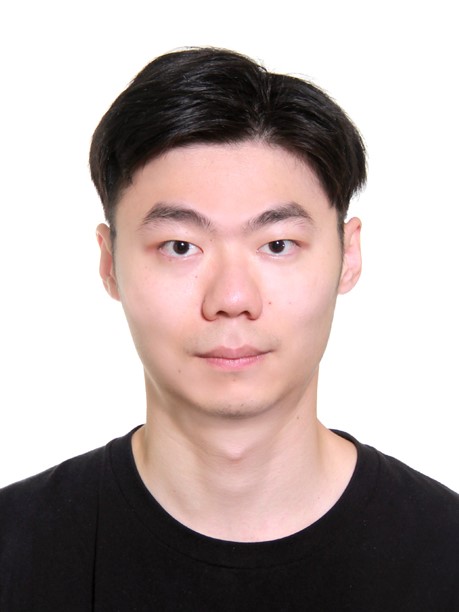}}]{Jingyuan Zhou} received the B.Eng. degree in Electronic Information Science and Technology from Sun Yat-sen University, Guangzhou, China, in 2022. He is currently working towards a Ph.D. degree with the National University of Singapore. His research interests include safety-critical control and privacy computing of mixed-autonomy traffic.
\end{IEEEbiography}
\vspace{-1cm}
\begin{IEEEbiography}
[{\includegraphics[width=1in,height=1.25in,clip,keepaspectratio]{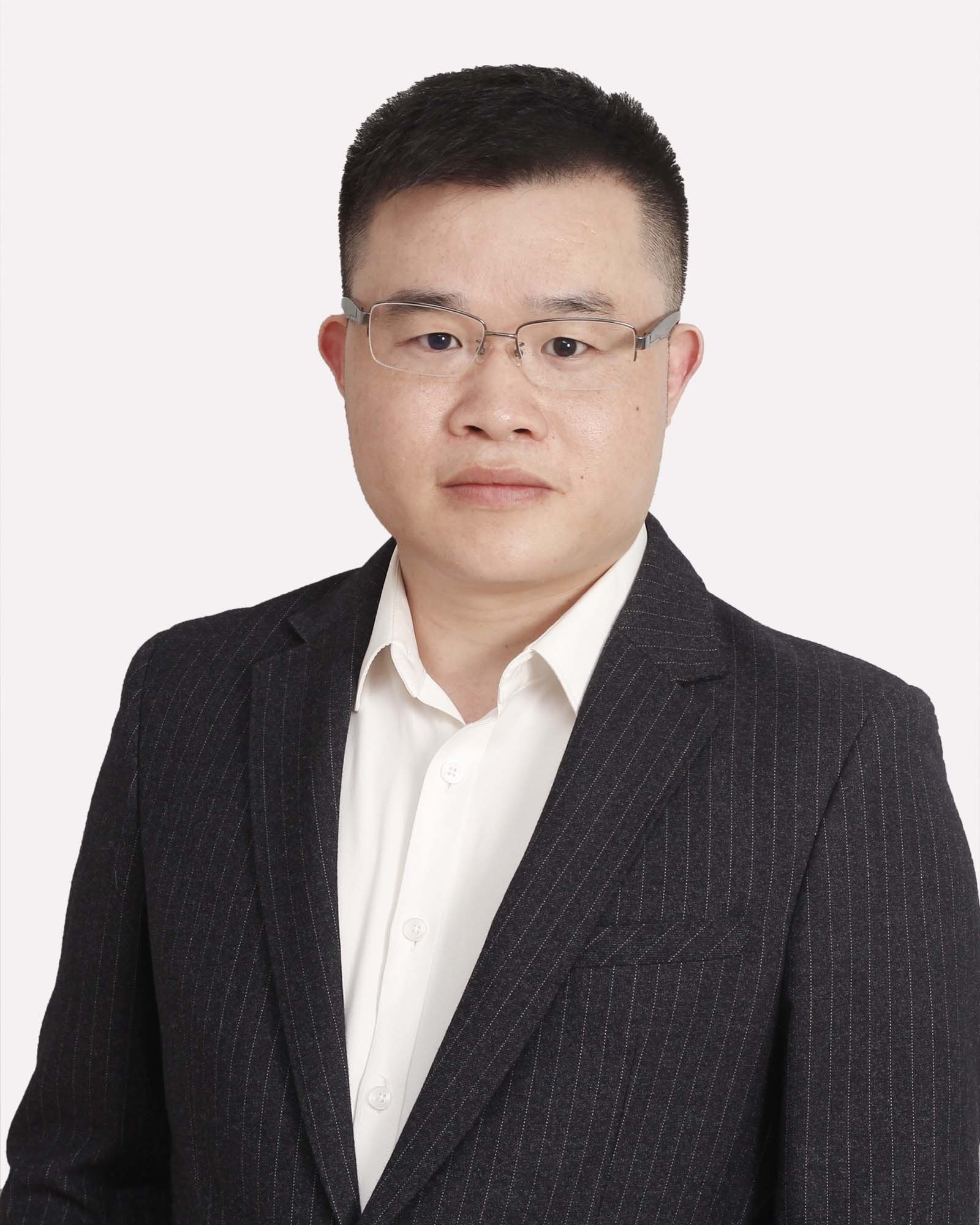}}]{Fu Zhang} 
received the B.E. degree in automation from the University of Science and
Technology of China (USTC), Hefei, China, in 2011 and the Ph.D. degree in controls from the University of California at Berkeley, Berkeley, CA, USA, in
2015. He joined the Department of Mechanical Engineering, University of Hong Kong (HKU), Hong Kong, as an Assistant Professor in 2018. His current research interests include robotics and controls, with a focus on unmanned aerial vehicle (UAV) design, navigation, control, and LiDAR-based SLAM.
\end{IEEEbiography}

\end{document}